\documentclass[5p]{elsarticle}

\usepackage{times}
\usepackage{epsfig}
\usepackage{balance}
\usepackage{latexsym}
\usepackage{amssymb}
\usepackage{enumerate}
\usepackage{graphicx}
\usepackage{subfigure}
\usepackage{multirow}
\usepackage{color}
\usepackage{amsmath}
\usepackage{algorithmic}
\usepackage{algorithm}
\usepackage{threeparttable}
\usepackage{braket}

\journal{Neurocomputing}

\begin{document}

\begin{frontmatter}

\title{Identifying The Most Informative Features Using A Structurally Interacting Elastic Net}


%
%

\author{Lixin Cui${}^{1}$\corref{cor1}}
\author{Lu Bai${}^{1}$ \corref{cor1}}
\author{Zhihong Zhang${}^{2}$}
\author{Yue Wang${}^{2}$}
\author{Edwin R. Hancock${}^{3}$}

\address{
${}^{1}$ Central University of Finance and Economics, Beijing, China.\\
${}^{2}$ Xiamen University, Xiamen, Fujian, China.\\
${}^{3}$ University of York, York, UK.
}
\cortext[cor1]{These authors contribute equally to this work and are co-first authors. Correspondence author: Lu Bai, email: bailucs@cufe.edu.cn.}

\begin{abstract}
Feature selection can efficiently identify the most informative features with respect to the target feature used in training. However, state-of-the-art vector-based methods are unable to encapsulate the relationships between feature samples into the feature selection process, thus leading to significant information loss. To address this problem, we propose a new graph-based structurally interacting elastic net method for feature selection. Specifically, we commence by constructing feature graphs that can incorporate pairwise relationship between samples. With the feature graphs to hand, we propose a new information theoretic criterion to measure the joint relevance of different pairwise feature combinations with respect to the target feature graph representation. This measure is used to obtain a structural interaction matrix where the elements represent the proposed information theoretic measure between feature pairs. We then formulate a new optimization model through the combination of the structural interaction matrix and an elastic net regression model for the feature subset selection problem. This allows us to a) preserve the information of the original vectorial space, b) remedy the information loss of the original feature space caused by using graph representation, and c) promote a sparse solution and also encourage correlated features to be selected. Because the proposed optimization problem is non-convex, we develop an efficient alternating direction multiplier method (ADMM) to locate the optimal solutions. Extensive experiments on various datasets demonstrate the effectiveness of the proposed method.
\end{abstract}


\end{frontmatter}


\section{Introduction}\label{s1}
There has recently been a rapid growth in both the size and dimension of the data encountered in many real world applications of pattern recognition including image processing, bioinformatics, and financial analysis. Finding useful information and building effective prediction models from such data presents new challenges for machine learning and pattern recognition~\cite{DBLP:journals/kbs/HanSH15}. One way to overcome this problem is to develop efficient spectral methods including stochastic neighbour embedding~\cite{DBLP:journals/ijon/LeePV15}, elastic embedding methods~\cite{DBLP:conf/icml/Carreira-Perpinan10} and feature selection~\cite{DBLP:journals/tkde/BenabdeslemH14} methods to reduce the dimensionality of the data.

Feature selection aims to identify an optimal subset of the most informative features by removing irrelevant and redundant features~\cite{DBLP:journals/tkde/BenabdeslemH14}. One of the main advantages is that feature selection can improve the predictive accuracy and enhance comprehensibility of learning tasks~\cite{DBLP:journals/kbs/WangSHQQ16}. Unlike feature extraction, feature selection can maintain the properties of the original features and has better interpretability. This is very important for understanding which features are most informative with respect to the target feature used in training. For instance, in Peer-to-Peer (P2P) lending analysis, it is important to understand which features of the P2P lending platforms (e.g., operation time, registered capital, and management team) affect the investors' decisions~\cite{DBLP:journals/eswa/Malekipirbazari15}. For medical diagnosis, it is crucial to know which characteristics of the patients (e.g., age, gender, and weight) affect the occurrence of a certain disease~\cite{DBLP:journals/cbm/ZhengL11}.

Because of these advantages, many efficient feature selection methods have been developed~\cite{DBLP:journals/kbs/WangSHQQ16}~\cite{DBLP:journals/tcyb/HouNLYW14}. Existing feature selection algorithms can be broadly categorized as filter and wrapper methods depending on whether the learning algorithm is used in the feature subset selection process~\cite{DBLP:journals/pami/NaghibiHP15}. Filter methods utilize the intrinsic properties of the data to build quantitative evaluation criteria~\cite{DBLP:journals/kbs/ElAlami09}. By contrast, wrapper methods ~\cite{DBLP:journals/kbs/WangACLA15} evaluate the selected feature subsets based on the performance measures of the classifier including accuracy and precision. Although wrapper methods often perform better than filter methods, they require significantly higher computational costs. In addition, in the presence of redundant features, wrappers tend to locate suboptimal subsets, and the characteristics of the selected subset are inevitably biased depending on the choice of the classifier~\cite{DBLP:journals/pami/NaghibiHP15}. Therefore, for high-dimensional data, filter methods are often preferred~\cite{DBLP:journals/ijon/QianS15}.

To construct effective filter methods, a good evaluation criterion is necessary. To date, many evaluation criteria have been used, including correlation~\cite{DBLP:conf/icml/Hall00}, consistency~\cite{DBLP:journals/ai/DashL03}, Fisher score \cite{DBLP:conf/nips/HeCN05}, and mutual information (MI) ~\cite{DBLP:journals/pr/HermanZ0YC13}. For instance, MI measures the mutual dependence of two variables~\cite{DBLP:journals/pr/HermanZ0YC13} and has been shown to have similar or better performance than other more sophisticated methods~\cite{DBLP:journals/jmlr/Fleuret04}. Due to its excellent performance, MI has received considerable attention for developing various information theoretic feature selection methods. Representative examples include 1) the Mutual Information-Based Feature Selection (MIFS)~\cite{DBLP:journals/tnn/Battiti94}, 2) the MIFS method under the assumption of a uniform distribution of input variables (MIFS-U)~\cite{DBLP:journals/pami/KwakC02}, 3) the Maximum-Relevance Minimum-Redundancy criterion (MRMR)~\cite{DBLP:journals/pami/PengLD05}, and 4) the Normalized Feature-Feature Mutual Information method (NMIFS)~\cite{DBLP:journals/tnn/EstevezTPZ09}. Although the performance of MI-based feature selection methods have been demonstrated in many applications, they suffer from two widely acknowledged limitations. First, they require the number of selected features to be predetermined. Second, they adopt greedy search methods to identify the most informative feature subsets~\cite{DBLP:journals/jmlr/Brown09}. To overcome these shortcomings, Liu et al.~\cite{DBLP:conf/aaai/LiuLLYXL11} have proposed the adaptive MI-based feature selection method that automatically determines the number of most informative features, by maximizing the average pairwise informativeness. Zhang and Hancock~\cite{DBLP:journals/prl/ZhangH12} have developed a hypergraph based information theoretic feature selection method that can automatically determine the size of the most informative feature subset through dominant hypergraph clustering~\cite{DBLP:journals/pami/PavanP07}.

However, none of the aforementioned information theoretic feature selection methods can incorporate pairwise relationship between samples of each feature dimension. More specifically, assume a dataset with $N$ features denoted as $$\mathcal{X}=\{\bold f_{1},\ldots,\bold f_{i},\ldots,\bold f_{N}\},$$ and each feature $\bold f_{i}$ has $M$ samples as $$\bold f_{i}=(f_{i1},\ldots,f_{ia},\ldots,f_{ib},\ldots,f_{iM})^T.$$ Traditional information theoretic feature selection methods represent each feature $\bold f_{i}$ as a vector, and thus ignore the relationship between pairwise samples $f_{ia}$ and $f_{ib}$ in $\bold f_{i}$. This deficiency restricts the precision of the information theoretic measure between pairs of features. To address this drawback, Cui et al.~\cite{DBLP:conf/sspr/CuiBW0ZH16} have recently developed a new feature selection method using graph-based features. Specifically, they transform each feature vector into a graph structure that encapsulates pairwise relationship between samples. The most relevant vectorial features are located by selecting the graph-based features that are most similar to the graph-based target feature, in terms of the Jensen-Shannon divergence measure between the graphs. To adaptively determine the most relevant feature subset, Cui et al.~\cite{DBLP:conf/gbrpr/CuiJB0H17} have further developed a new information theoretic feature selection method which a) encapsulates the relationship between sample pairs for each feature dimension and b) automatically identifies the subset containing the most informative and least redundant features by solving a quadratic programming problem.

However, the aforementioned graph-based feature selection methods may lead to significant information loss concerning the relationships between samples from the original vector space. To illustrate this point, assume that two pairs of samples from the same feature dimension $\bold f_{i}$ are denoted as $\{f_{i1}, f_{i3}\}$ and $\{f_{i3},f_{i5}\}$, respectively. Following Cui et al.~\cite{DBLP:conf/gbrpr/CuiJB0H17}, we transform the feature vector $\bold f_{i}$ into a graph-based representation $\bold G_{i}$, which is a complete weighted graph. Each vertex $v_{a}$ of $\bold G_{i}$ represents a corresponding sample $f_{ia}$ in $\bold f_{i}$ and each weighted edge $(v_{a},v_{b})$ represents the relationship between sample pair $f_{ia}$ and $f_{ib}$. If the Euclidean distances of the two pairs, i.e., $\{f_{i1}, f_{i3}\}$ and $\{f_{i3},f_{i5}\}$ are the same, the weights associated with the two pairs of samples are also the same in the feature graph $\bold G_{i}$. However, these two pairs of samples are located differently in the original vector space. This means that the graph-based feature representation may lead to information loss. One exception is that the vertex label is the associated sample value of the original features, i.e., the vertex label is continuous. However, in this case, we need to measure the affinity between a pair of graphs associated with continuous vertex labels and this results in significantly higher computational complexity~\cite{DBLP:conf/nips/FeragenKPBB13}.

To summarize the above, it is fair to say that it still remains a challenge to develop an effective information theoretic feature selection methods that can both encapsulate pairwise relationship between samples of each feature dimension and avoid information loss from the original vector space.

On the other hand, sparse feature selection methods have received increasing attention~\cite{DBLP:journals/pr/YanY15}. By formulating feature selection as a regression model with an ordinary least square (OLS) term  and a specifically designed sparsity inducing regularizer, the regression model can be efficiently represented by a linear combination of a set of the most active variables. The cardinality of the set of the selected variables is significantly smaller than the entire number of variables~\cite{DBLP:journals/prl/ZhangTBXH17}. In other words, the regression model retains information concerning the original feature space and also allows us to adaptively select the most informative feature subset. Because of these advantages, many efficient regularization techniques including Lasso~\cite{Lasso}, Elastic Net~\cite{ElasticNet}, and Group Lasso~\cite{DBLP:journals/bmcbi/MaSH07} have been extensively studied for high-dimensional data feature selection. For instance, Zheng and Liu~\cite{DBLP:journals/cbm/ZhengL11} have developed a Lasso operator to identify the most informative features for cancer classification, where Lasso enforces automatic feature selection by forcing at least some features to zero. Panagakis et al.~\cite{DBLP:journals/prl/PanagakisK14} have developed a new similarity measure based on the matrix Elastic Net regularization to efficiently deal with highly correlated audio feature vectors. Marafino et al.~\cite{DBLP:journals/jbi/MarafinoBD15} have proposed an efficient sparse feature selection method for biomedical text classification using the Elastic Net. More recently, Zhang et al.~\cite{DBLP:journals/prl/ZhangTBXH17} have devised a new regularization term in the Lasso regression model to impose high order interactions between covariates and responses. The high-order relations among covariates are represented by a feature hypergraph and then used as a regularizer on the covariate coefficients to automatically select the most relevant features.

Although sparsity is desirable for designing effective feature selection algorithms, it is worth noting that most of the existing sparse feature selection methods seldom consider pairwise relationship between samples from each feature dimension. Intuitively, such structural information is important for improving the efficiency of feature selection methods. In addition, as opposed to the Elastic Net, the use of Lasso proves to be problematic when at least some features are highly correlated. In this case, Lasso selects one feature at random. As a result, given $n$ training samples, Lasso can only select at most $n$ features.

Motivated by the above discussion, we aim to overcome the limitations of existing feature selection methods by developing a novel structurally interacting elastic net feature selection method. The proposed method not only considers the structural relationships between feature samples but also remedies the information loss caused by the graph-based representation of features. In addition, we also explore how to ensure sparsity and promote a grouping effect among selected features via elastic net regularization.

\subsection{Related Work}
Feature selection has been widely studied in machine learning and pattern analysis. The topic of feature selection has been reviewed in a number of recent papers~\cite{DBLP:journals/jmlr/BrownPZL12},~\cite{DBLP:journals/nca/VergaraE14} and~\cite{DBLP:journals/kbs/Bolon-CanedoSA15}. In this section, we briefly state-of-the-art MI-based and sparse feature selection methods, which are related to our proposed method.

MI is often considered as an evaluation criterion to measure the relevance between features and the target labels due to its effectiveness at quantifying how much information is shared by two random variables. Because of this, MI has been extensively used for developing information theoretic feature selection methods. In the earlier reported work, Battiti~\cite{DBLP:journals/tnn/Battiti94} introduced a first order incremental search algorithm based on MI known as the MIFS criterion,

\begin{equation}
J_{MIFS}=I(f_{i};C)-\beta\sum_{f_{s}\in S}^{}I(f_{s};f_{i}).
\end{equation}

For a given set of existing selected features $S$, at each step MIFS locates the candidate feature $f_{i}$ which maximizes the relevance to the class $I(f_{i};C)$, instead of calculating the joint MI between the selected features and the class label $C$. The proportional term $\beta I(f_{s};f_{i})$ measures the overlap information between the candidate feature and existing features, and is used to regulate the feature selection process. The parameter $\beta$ may significantly influence the features selected and needs to be carefully controlled. It is worth noting that because MIFS only considers features that have maximum MI with the output classes, it might treat features that have rich information about the output class as redundant, leading to a suboptimal subset. To overcome this drawback, Kwak and Choi~\cite{DBLP:journals/pami/KwakC02} improved MIFS by developing MIFS-U under the assumption of a uniform distribution for selected features. This uniform criterion is defined as
\begin{equation}
J_{MIFS-U}=I(f_{i};C)-\beta\sum_{f_{s}\in S}\frac{I(f_{s};C)}{H(f_{s})}I(f_{i};f_{s}),
\end{equation}
where $H(f_{s})=-\sum_{f_{s}\in S} P(f_{s} logP(f_{s}))$ is the entropy associated with the probability distribution for $f_{s}$. Instead of calculating $I(C;f_{i}|S)$ directly, only $I(S;f_{i})$ and $I(C;f_{i})$ are computed. The conditional MI (denoted as $I(C;f_{i}|S)$) between the class label $C$ and the candidate feature $f_{i}$ for a given feature subset $S$ can be approximated as
$I(C;f_{i}|S)=I(C;f_{i})-{I(S;f_{i})-I(S;f_{i}|C)}$.

Although MIFS-U gives better estimation than MIFS, the model parameters need to be carefully controlled to avoid bad results. To overcome this problem, Peng et al.~\cite{DBLP:journals/pami/PengLD05} proposed a parameter-free method, referred to as MRMR, which is defined as
\begin{equation}
J_{MRMR}=I(f_{i};C)-\frac{1}{|S|}\sum_{f_{s}\in S}I(f_{i};f_{s}),
\end{equation}
where $|S|$ is the cardinality of the selected feature set $S$. MRMR uses the average of the redundancy term to eliminate the difficulty of parameter selection of $\beta$ with MIFS and MIFS-U methods. However, as a first-order incremental method, that sequentially selects one feature after another based on the evaluation function, MRMR presents similar limitations as MIFS and MIFS-U in the presence of many irrelevant or redundant features. This is because the conditional MI $I(C;f_{i}|S)$ between the target class $C$ and the candidate feature $f_{i}$ for a given subset of features $S$ is ignored. To deal with this problem, Estevez et al.~\cite{DBLP:journals/tnn/EstevezTPZ09} developed the normalized NMIFS method to achieve a balance between the relevance and the redundancy term, defined as
\begin{equation}
J_{NMIFS}=I(f_{i};C)-\frac{1}{|S|}\sum_{f_{s}\in S}\hat{I}(f_{i};f_{s}),
\end{equation}
where $\hat{I}(f_{i};f_{s})=\frac{I(f_{i}f_{s})}{\inf(H(f_{s}),H(f_{i}))}$ is the normalized MI.

On the other hand, sparse feature selection methods have recently attracted much attention. Typically we have a set of training data $\{(x_{i},y_{i}),i=1,...,N\}$, which is used to estimate the regression coefficients $\beta$. Each $x_{i}=(f_{1}^{i},f_{2}^{i},...,f_{d}^{i})^T\in R^{d\times1}$ is a predictive vector of feature measurements for the ith sample. To fit the linear regression model, the most popular ordinary least square (OLS) method is adopted. OLS selects the coefficients $\beta=(\beta_{1},...,\beta_{d})^T$ by minimizing the residual sum of squares denoted as
\begin{align}
&\min_{\beta\in\Re^{d}}\sum_{i=1}^{N}\|y_{i}-\sum_{j=1}^{d}\beta_{j}f_{j}^{i}\|_{2}^{2}=\min_{\beta\in\Re^{d}}\|\bold y^{T}-\beta^{t}\bold X\|_{2}^{2}\nonumber\\
&s.t.\sum_{i=1}^{N}\|\beta\|_{0}=k\label{OLS},
\end{align}
where $\bold y \in \Re^{N\times1}$ is the label (response) vector, $\bold X \in \Re^{d\times N}$ is the training dataset, $k$ is a predetermined number of selected features. The minimisation of Eq.(\ref{OLS}) has been proved to be a NP-hard optimization problem and is very difficult to be solved. In practice, we can relax the constraint equation by imposing a positive regularization parameter $\lambda$ and adding it to the objective function, that is
\begin{equation}
\min_{\beta\in\Re^{d}}\|\bold y^{T}-\beta^{t}\bold X\|_{2}^{2}+\lambda\|\beta\|_{0}\label{OLS1}.
\end{equation}

Unfortunately solving Eq.(\ref{OLS1}) is still challenging. Therefore, an alternative formulation using $l_{1}$-norm regularization instead of $l_{0}$-norm has been proposed for practical purposes. This corresponds to the regularized counterpart of the Lasso (least absolute shrinkage and selection operator) problem in statistical learning~\cite{Lasso}. Lasso imposes a $l_{1}$ constraint on the regression coefficients, so that some of the regression coefficients in the regression model will shrink to zeros. Thus it can automatically select a set of the informative variables. Correspondingly, the feature selection problem with Lasso penalty is defined as
\begin{equation}
\min_{\beta\in\Re^{d}}\|\bold y^{T}-\beta^{t}\bold X\|_{2}^{2}+\lambda\|\beta\|_{1}\label{Lasso},
\end{equation}
where $\|\beta_{1}\|$ is the $l_{1}$-norm of vector $\beta$, that is, $\|\beta_{1}\|=\sum_{j=1}^{d}|\beta_{j}|$. The parameter $\lambda\geq0$ controls the amount of regularization applied to the estimate. The larger $\lambda$, the larger the number of zeros in $\beta$. The nonzero components give the selected variables. After the optimal value of $\beta$ is obtained, one can choose the feature indices corresponding to the top $k$ largest values of the summation of the absolute values among each column.

Lasso requires the independence assumption of the input variables, however, in most real world data, features are often correlated. Therefore, in the presence of highly correlated features, Lasso tends to select only one of these features at random, resulting in suboptimal performance (37). Moreover, the $l_{1}$ minimization algorithm is not stable when compared with $l_{2}$ minimization (30). For this reason, the elastic net (5) adds an additional $l_{2}$ regularization term into the Lasso objective function, which can be formulated as
\begin{equation}
\min_{\beta}\|\bold y^{T}-\beta^{t}\bold X\|_{2}^{2}+\lambda_{1}\|\beta\|_{1}+\lambda_{2}\|\beta\|_{2}^{2}\label{EN},
\end{equation}
where $\lambda_{1},\lambda_{2}\geq 0$ are the tuning parameters.

Elastic Net can be seen as a linear combination of the Lasso and Ridge penalty. When $\lambda_{1}=0$, it becomes simple Ridge regression, when $\lambda_{2}=0$, it is equivalent to the Lasso penalty. Thus, Elastic Net enjoys a similar sparsity of representation of Lasso and also allows groups of correlated features to be selected. In the literature, it is reported that Elastic Net usually outperforms Lasso when the number of features is much larger than the number of samples~\cite{ElasticNet}.

In summary, most of the exiting MI-based feature selection methods aim to develop an efficient quantitative evaluation criterion that simultaneously maximizes relevancy and minimizes redundancy. Unfortunately, it has been noted that such MI-based feature selection methods have two common limitations. First, they tend to ignore pairwise relationship between samples of each feature dimension, which leads to significant information loss. Second, the majority of these methods cannot adaptively identify the most informative feature subset. Alternatively, sparse feature selection methods like Lasso and Elastic Net ensure parameter vector sparsity and allow the relevant features to be adaptively selected. However, existing sparse feature selection methods also fail to encapsulate pairwise relationship between samples for each feature dimension. These drawbacks motivate us to develop a novel structurally interacting elastic net feature selection method to adaptively locate the most informative feature subset.

\subsection{Contributions}
As previously stated, the aim of this paper is to overcome the limitations of existing MI-based and sparse feature selection methods by developing a new structurally interacting elastic net feature selection algorithm. In summary, the contributions of this work are threefold.
First, we transform each vector feature into a graph-based feature representation, where each vertex represents a corresponding sample in each feature dimension and each weighted edge represents the pairwise relationship between samples from each feature dimension. We use the Euclidean distance to measure the pairwise relationship between samples. Similarly, we also construct a complete feature graph for the target feature. To measure the joint relevance of different pairwise feature combinations with respect to the target feature, we propose a new information theoretic criterion using the Jensen-Shannon divergence (JSD). Based on this criteria, we obtain a new interaction matrix which characterizes the informative relationships between feature pairs.
Second, to a) incorporate the pairwise sample relationships in each feature dimension, b) remedy information loss in the original feature space, and c) adaptively select the most informative feature subset, we formulate the proposed graph-based feature selection method as an elastic net regression model. Specifically, the interaction matrix encapsulates high-dimensional structural relationships between feature samples and thus provides richer representation of structural interaction information between features. The ordinary least-square (OLS) term utilizes information from the original feature space, and thus remedies the information loss caused by representing features as graphs. In addition, the $l_{1}$-norm regularizer ensures sparsity in the coefficients of variables and the $l_{2}$-norm regularizer promotes a grouping effect to select correlated features.
Third, an efficient alternating direction multiplier method (ADMM) is presented to solve the proposed elastic net optimization problem. Comprehensive experiments on eight standard machine learning datasets and two publically available datasets demonstrate the effectiveness of the proposed method.

\subsection{Paper Outline}
The remainder of the paper is organized as follows. Section~\ref{s2} discusses the important concepts which will be used for the proposed feature selection method. Section~\ref{s3}  presents the proposed structurally interacting elastic net for feature selection. Section~\ref{s4} provides our experimental evaluation. Finally, Section~\ref{s5} concludes our work.

\section{Preliminary Concepts}\label{s2}
In this section, we review some preliminary concepts that will be used in this work. We commence by reviewing how to construct the feature graph to incorporate structural information for feature samples. Then we introduce the concept of Jensen-Shannon divergence, which will be used to calculate the similarity between feature graphs.
\subsection{Construction of Feature Graphs}\label{s2.2}
In this subsection, we introduce how to transform each vectorial feature into a complete weighted graph. The advantages of using the graph-based representation are twofold. First, graph structures have a stronger ability to encapsulate global topological information than vectors. Second, graphs can incorporate the relationships between samples of each original vector feature into the feature selection process, thus reducing information loss.

Given a dataset of $N$ features denoted as $$\mathcal{X}=\{\bold f_{1},\ldots,\bold f_{i},\ldots,\bold f_{N}\}\in \mathbb{R}^{M\times N}$$, $\bold f_{i}$ represents the $i$-th vectorial feature and has $M$ samples $\bold f_{i}=(f_{i1},\ldots,f_{ia},\ldots,f_{ib},\ldots,f_{iM})^T$. We transform each feature $\bold f_{i}$ into a graph-based feature representation $\bold G_{i}(V_i,E_i)$, where the vertex $v_{ia}\in V_i$ indicates the $a$-th sample $f_{ia}$ of $\bold f_{i}$. Each pair of vertices $v_{ia}$ and $v_{ib}$ are connected by a weighted edge $(v_{ia},v_{ib})\in E_i$, and the dissimilarity weight $\omega(v_{ia},v_{ib})$ of $(v_{ia},v_{ib})$ is the Euclidean distance between $f_{ia}$ and $f_{ib}$, i.e.,

\begin{equation}
\omega(v_{ia},v_{ib})= \sqrt{(f_{ia} -  f_{ib})^{2}} \label{GBF}.
\end{equation}

Similarly, if the sample values of the target feature $\bold Y=(y_{1},\ldots,y_{a},\ldots,y_{b},\ldots,y_{M})^T$ are continuous, its graph-based feature representation $\bold{\hat{G}}(\hat{V},\hat{E})$ can be computed using Eq.(\ref{GBF}) and the vertex $\hat{v}_a$ represents the $a$-th sample $y_{a}$. However, for classification problems, the target features $Y$ are the class labels and thus takes the discrete values $c\in\{1,2,\ldots,C\}$, i.e., the samples of each feature $\bold f_{i}$ are assigned to the $C$ different classes. In this case, we propose to compute the graph-based target feature $\bold{\hat{G}_{i}}(\hat{V}_{i},\hat{E}_{i})$ for each feature $\bold f_{i}$, where the dissimilarity weight $\omega(\hat{v}_{ia},\hat{v}_{ib})$ of each edge $(\hat{v}_{ia},\hat{v}_{ib})\in \hat{E}_i$ is
\begin{equation}
\omega(\hat{v}_{ia},\hat{v}_{ib})= \sqrt{(\mu_{ia} -  \mu_{ib})^{2}},\label{GBTFD}
\end{equation}
where $\mu_{ia}$ is the mean value of all samples in $\bold f_{i}$ from the same class $c$.


\subsection{The Jensen-Shannon Divergence}
In information theory, the JSD is a dissimilarity measure between probability distributions. Let two (discrete) probability distributions be $\mathcal{P}=(p_1,\ldots,p_a,\ldots,p_A)$ and
$\mathcal{Q}=(q_1,\ldots,q_b,\ldots,q_B)$, then the JSD between $\mathcal{P}$ and $\mathcal{Q}$ is defined as
\begin{equation}\label{Eq:CJSD}
\mathrm{JSD}(\mathcal{P},\mathcal{Q}) = H_S\Big(\frac{\mathcal{P}+\mathcal{Q}}{2}\Big) - \frac{1}{2} H_S(\mathcal{P}) -
\frac{1}{2} H_S(\mathcal{Q}),
\end{equation}
where $H_S(\mathcal{P})=\sum_{a=1}^A p_a \log p_a$ is the Shannon entropy of the probability distribution $\mathcal{P}$. In~\cite{DBLP:conf/pkdd/Bai0BH14}, the JSD has been used as a means of measuring the information theoretic dissimilarity between graphs associated with their probability distributions. In this work, we are concerned with the similarity between graph-based feature representations. Therefore, we adopt the negative exponential of $\mathrm{JSD}(\mathcal{P},\mathcal{Q})$
to compute the similarity measure $I_S$ between probability distributions, i.e.,
\begin{equation}\label{Eq:ExpoentialJSD}
I_S(\mathcal{P},\mathcal{Q})=\exp\{-\mathrm{JSD}(\mathcal{P},\mathcal{Q})\}.
\end{equation}

\section{The Proposed Feature Selection Method}\label{s3}
In this section, we introduce our proposed structurally interacting elastic net feature selection method. We first detail the formulation of the structurally interacting elastic net model. To solve the optimization model, the alternating direction method of multiplier (ADMM) algorithm~\cite{DBLP:journals/ftml/BoydPCPE11} is used to identify the most informative feature subset. Finally, we provide the convergence proof and complexity analysis for the method.

We propose to use the following information theoretic criterion to measure the joint relevance of different pairwise feature combinations with respect to target labels. For a set of $N$ features $\bold f_{1},\ldots,\bold f_{i},\ldots,\bold f_{j},\ldots,\bold f_{N}$ and the associated continuous target feature $\bold Y$, the relevance degree of the feature pair $\{\bold f_{i},\bold f_{j}\}$ is
\begin{equation}\label{ITC}
W_{\bold f_{i},\bold f_{j}}=\frac{I_S(\bold G_{i},\bold{\hat{G}})+I_{S}(\bold G_{j},\bold{\hat{G}})}{I_{S}(\bold G_{i},\bold G_{j})},
\end{equation}
where $\bold G_{i}$ and $\bold{\hat{G}}$ are the graph-based feature representations of $\bold f_{i}$ and $\bold Y$, $I_{S}$ is the JSD based information theoretic similarity measure defined in Eq.(\ref{Eq:ExpoentialJSD}). The above relevance measure consists of three terms. The first and second terms $I_S(\bold G_{i},\bold{\hat{G}})$ and $I_{S}(\bold G_{j},\bold{\hat{G}})$ are the relevance degrees of individual features $\bold f_{i}$ and $\bold f_{j}$ with respect to the target feature $\bold Y$, respectively. The third term $I_{S}(\bold G_{i},\bold G_{j})$ measures the relevance between the feature pair $\{\bold f_{i},\bold f_{i}\}$. Therefore, $W_{\bold f_{i},\bold f_{j}}$ is large if and only if both $I_S(\bold G_{i},\bold{\hat{G}})$ and $I_{S}(\bold G_{j},\bold{\hat{G}})$ are large (i.e., both $\bold f_{i}$ and $\bold f_{j}$ are informative themselves with respect to the target feature representation $\bold Y$) and $I_{S}(\bold G_{i},\bold G_{j})$ is small (i.e., $\bold f_{i}$ and $\bold f_{j}$ are not relevant).

For classification problems, the samples of the target feature $\bold Y$ take the discrete value $c$ and $c\in\{1,2,\ldots,C\}$. In this case, we compute the individual graph-based target feature representation $\bold{\hat{G}_{i}}$ for each feature $\bold f_{i}$, and the relevance measure defined in Eq.(\ref{ITC}) can be written as
\begin{equation}\label{ITD}
W_{\bold f_{i},\bold f_{j}}=\frac{I_S(\bold G_{i},\bold{\hat{G}}_i)+I_{S}(\bold G_{j},\bold{\hat{G}}_j)}{I_{S}(\bold G_{i},\bold G_{j})}.
\end{equation}

Similarly to Eq.(\ref{ITC}), the three terms of Eq.(\ref{ITD}) have the same corresponding theoretical significance.

Furthermore, based on the graph-based feature representations, we construct a feature informativeness matrix $\mathbf{W}$, where each element $W_{i,j}\in \mathbf{W}$ represents the information theoretic measure between a feature pair $\{\bold f_i,\bold f_j\}$ based on Eq.(\ref{ITC}) (for $\bold Y$ is continuous) or Eq.(\ref{ITD}) (for $\bold Y$ is discrete). Given the informativeness matrix $\mathbf{W}$ and the d-dimensional feature indicator vector $\bold \beta$, where $\beta_i$ represents the coefficient for the i-th feature, we can identify the informative feature subset by solving the following maximization problem
\begin{equation}\label{SINT}
\max f(\beta)= \max_{\beta\in \Re^{d}} \beta^{T}\bold W \beta,
\end{equation}
subject to $\mathbf{\beta} \in \mathbb{R}^{N}$, $\mathbf{\beta} \geq 0$. The solution vector $$\mathbf{\beta}=(\beta_{1},...,\beta_{d})^{T}$$ to the above quadratic program is an $N$-dimensional vector. When $\beta_{i}>0$, the $i$-th feature $\bold f_i$ belongs to the most informative feature subset i.e., feature $\bold f_{i}$ is selected if and only if the $i$-th component of $\mathbf{\beta}$ is positive ($i\in\{1,2,...,d\}$).

\subsection{The Proposed Structurally Interacting Elastic Net}
The proposed feature subset selection algorithm aims to incorporate structural information between pairwise features and simultaneously allow correlated features to be selected, as well as promote a sparse solution. Therefore, we combine Eq.(\ref{SINT}) and Eq.(\ref{EN}) to construct the associated structurally interacting elastic net for feature selection with the following mathematical form
\begin{equation}\label{INT1}
\min_{\beta\in \Re^{d}}\frac{1}{2}\|\mathbf{y}^{T}-\beta^{T}\mathbf{X}\|^{2}_{2}+\lambda_{1}\|\beta\|_{1}+\lambda_{2}\|\beta\|^{2}_{2}-\lambda_{3}\beta^{T}\bold W\beta,
\end{equation}
where $\lambda_{1}$ and $\lambda_{2}$ are the tuning parameters in the elastic net regression model, and $\lambda_{3}$ is the associated tuning parameter for the structural interaction matrix $\bold W$.

It can be seen that the first term in Eq.(\ref{INT1}) remedies the information loss from the original feature space, while the second and third terms ensure the sparsity and grouping among selected features. The fourth term incorporates structural information concerning the relationships between feature samples. Because $\beta^{T} \bold W \beta$ is a non-convex constraint, the proposed method distinguishes itself from existing Lasso-type methods using convex optimization methods, which may become trapped in suboptimal solutions. Specifically, for the proposed model (\ref{INT1}), we need to develop efficient algorithms to obtain the optimal solutions (denoted as $\beta^{\ast}$). A feature $f_{i}$ is selected if and only if $\beta_{i}^{\ast}>0$. Consequently, we can recover the number of features in the optimal feature subset according to the number of positive components of $\beta^{\ast}$.

\subsection{Optimization Algorithm}
To solve the non-convex problem (\ref{INT1}), we develop an optimization algorithm using ADMM~\cite{DBLP:journals/ftml/BoydPCPE11}. The ADMM approach is a powerful algorithm that is well suited to problems arising in machine learning. The basic principle of the ADMM approach is to decompose a hard optimization problem into a series of smaller ones, each of which is simpler to handle. It takes the form of a decomposition-coordination procedure, in which the solutions to small local subproblems are coordinated to find a solution to a large global problem. ADMM can be viewed as an attempt to blend the benefits of dual decomposition and augmented Lagrangian methods for constrained optimization. It turns out to be equivalent or closely related to many well known algorithms as well, such as Douglas-Rachford splitting from numerical analysis~\cite{DBLP:journals/tac/GiselssonB17}, proximal methods~\cite{DBLP:journals/jota/BentoFM17}, and many others.

In ADMM form, problem (\ref{INT1}) can be re-written as
\begin{align}
&\min_{\beta\in \Re^{d}}\frac{1}{2}\|\mathbf{y}^{T}-\beta^{T}\mathbf{X}\|^{2}_{2}+\lambda_{2}\|\beta\|^{2}_{2}-\lambda_{3}\beta^{T}\bold W\beta+\lambda_{1}\|\gamma\|_{1}\nonumber\\
&s.t.\quad \beta-\gamma=0,
\end{align}
where $\gamma$ is an auxiliary variable, which can be regarded as a proxy for vector $\beta$. By doing so, the objective function can be split into two separate parts associated with two different variables, i.e., $\beta$ and $\gamma$, indicating that the hard constrained problem can be solved separately. As in the method of multipliers, we form the augmented Lagrangian function associated with the constrained problem (\ref{INT1}) as follows
\begin{align}
L_{\rho}(\beta,\gamma,z)=\frac{1}{2}\|\mathbf{y}^{T}-\beta^{T}\mathbf{X}\|^{2}_{2}+\lambda_{2}\|\beta\|^{2}_{2}-\lambda_{3}\beta^{T}\bold W\beta \nonumber \\
+\lambda_{1}\|\gamma\|_{1}+<\beta-\gamma,z>+\frac{\rho}{2}\|\beta-\gamma\|^{2}_{2},
\end{align}
where $\langle\cdot,\cdot\rangle$ is an Euclidean inner product, $z$ is a dual variable (i.e.,the Lagrange multiplier) associated with the equality constraint $\beta=\gamma$, and $\rho$ is a positive penalty parameter (step size for dual variable update). By introducing an additional variable $\gamma$ and an additional constraint $\beta-\gamma=0$, we have simplified the optimization problem (\ref{INT1}) by decoupling the objective function into two parts that depend on two different variables. In other words, ADMM decomposes the minimization of $L_{\rho}(\beta,\gamma,z)$ into two simpler subproblems. Specifically, ADMM solves the original problem (\ref{INT1}) by seeking for a saddle point of the augmented Lagrangian by iteratively minimizing $L_{\rho}(\beta,\gamma,z)$ over $\beta$, $\gamma$, and $z$. In ADMM, the variables $\beta$, $\gamma$, and $z$ are updated in an alternating or sequential fashion, which accounts for the term alternating direction. This updating rule is shown below

(1) $\beta^{k+1}=\arg\min_{\beta\in \Re^{d}}L(\beta,\gamma^{k},z^{k})$, //$\beta$-minimization

(2) $\gamma^{k+1}=\arg\min_{\beta\in \Re^{d}}L(\beta^{k+1},\gamma,z^{k})$, //$\gamma$-minimization

(3) $z^{k+1}=z^{k}+\rho(\beta^{k+1}-\gamma^{k+1})$, //$z$-update

Given the above updating rule, we need to resolve each sub-problem iteratively until the termination criteria is satisfied. Using ADMM, we perform the following calculation steps at each iteration.

(a)\textbf{Update} \textbf{$\beta$}

In the $(k+1)-th$ iteration, in order to update $\beta^{k}$, we need to solve the following sub-problem, where the values of $\gamma^{k}$ and $z^{k}$ are fixed
\begin{align}
\min_{\beta\in \Re^{d}}\frac{1}{2}\|\mathbf{y}^{T}-\beta^{T}\mathbf{X}\|^{2}_{2}+\lambda_{2}\|\beta\|^{2}_{2}-\lambda_{3}\beta^{T}\bold W\beta+\lambda_{1}\|\gamma\|_{1} \nonumber \\
+<\beta-\gamma^{k},z^{k}>+\frac{\rho}{2}\|\beta-\gamma^{k}\|^{2}_{2}.
\end{align}

Let the partial derivative with respect to $\beta$ be equal to zero, we have
\begin{align}
\partial[{\min_{\beta\in \Re^{d}}\frac{1}{2}\|\mathbf{y}^{T}-\beta^{T}\mathbf{X}\|^{2}_{2}+\lambda_{2}\|\beta\|^{2}_{2}-\lambda_{3}\beta^{T}\bold W\beta} \\ \nonumber+\lambda_{1}\|\gamma\|_{1}+<\beta-\gamma^{k},z^{k}>+\frac{\rho}{2}\|\beta-\gamma^{k}\|^{2}_{2}] / \partial\beta =0,
\end{align}

because
\begin{equation}
\left\{
\begin{array}{ll}
\frac{\partial(\frac{1}{2}\|\mathbf{y}^{T}-\beta^{T}\mathbf{X}\|^{2}_{2})}{\partial\beta}=-\mathbf{X}\mathbf{y}+\mathbf{X}\mathbf{X}^{T}\beta \\
\frac{\partial(\lambda_{2}\|\beta\|^{2}_{2})}{\partial\beta}=\lambda_{2}\beta \\
\frac{\partial(-\lambda_{3}\beta^{T}\bold W\beta)}{\partial\beta}=-2\lambda_{3}\bold W\beta \\
\frac{\partial<\beta-\gamma^{k},z^{k}>}{\partial\beta}=z^{k} \\
\frac{\partial(\frac{\rho}{2}\|\beta-\gamma^{k}\|^{2}_{2})}{\partial\beta}=\rho(\beta-\gamma^{k}),
\end{array}\right.
\label{ADMMBeta}
\end{equation}

we have
\begin{equation}
-\mathbf{X}\textbf{y}+\mathbf{X}\mathbf{X}^{T}\beta+\lambda_{2}\beta-2\lambda_{3}\bold W\beta+z^{k}+\rho(\beta-\gamma^{k})=0,
\end{equation}
that is,
\begin{equation}
\beta^{k+1}=(\mathbf{X}\mathbf{X}^{T}+\lambda_{2}\mathbf{I}-2\lambda_{3}\bold W+\rho \mathbf{I})^{-1}(\mathbf{X}\textbf{y}-z^{k}+\rho\gamma^{k})\label{ADMM2}.
\end{equation}

(b)\textbf{Update} \textbf{$\gamma$}

Based on the results, assume $\beta^{k+1}_{i}$ and $z^{k}_{i}$ are fixed, for $i=1,2,...,d$, we update $\gamma_{i}^{k+1}$ by solving the following sub-optimization problem
\begin{equation}
\min_{\gamma_{i}}\lambda_{1}\sum_{i=1}^{d}\|\gamma_{i}\|_{1}-\sum_{i=1}^{d}<\gamma_{i},z_{i}^{k}>+\frac{\rho}{2}\sum_{i=1}^{d}\|\beta_{i}^{k+1}-\gamma_{i}\|_{2}^{2},
\end{equation}

\begin{equation}
\frac{\partial[\min_{\gamma_{i}}\lambda_{1}\sum_{i=1}^{d}\|\gamma_{i}\|_{1}-\sum_{i=1}^{d}<\gamma_{i},z_{i}^{k}>+\frac{\rho}{2}\sum_{i=1}^{d}\|\beta_{i}^{k+1}-\gamma_{i}\|_{2}^{2}]}{\partial\gamma_{i}}=0.
\end{equation}
We therefore have the following results

\begin{equation}
\gamma_{i}^{k+1}=
\left\{
\begin{array}{ll}
\frac{1}{\rho}(z_{i}^{k}+\rho\beta_{i}^{k+1}-\lambda_{1}),& \textrm{if $z_{i}^{k}+\rho\beta_{i}^{k+1}>\lambda_{1}$} \\
\frac{1}{\rho}(z_{i}^{k}+\rho\beta_{i}^{k+1}-\lambda_{1}),& \textrm{if $z_{i}^{k}+\rho\beta_{i}^{k+1}<-\lambda_{1}$} \\
0, & \textrm{if $z_{i}^{k}+\rho\beta_{i}^{k+1}\in[-\lambda_{1},\lambda_{1}]$}
\end{array}\right.
\label{ADMM3}
\end{equation}

(c)\textbf{Update}  $\textbf{z}$

Then, assume $\beta^{k+1}_{i}$ and $\gamma^{k+1}_{i}$ are fixed, for $i=1,2,...,d$, we update $z_{i}^{k+1}$ by using the following equation
\begin{equation}
z_{i}^{k+1}=z_{i}^{k}+\rho(\beta_{i}^{k+1}-\gamma_{i}^{k+1}).\label{ADMM4}
\end{equation}

Based on procedures (a), (b), and (c), we summarize the optimization algorithm below

\begin{algorithm}
\small{
\caption{\small{The proposed ADMM algorithm for structurally interacting Elastic Net.}}
\textbf{Input}: $\mathbf{X},\mathbf{y},\beta^{0},z^{0},\lambda_{1},\lambda_{2},\lambda_{3},\rho$

\textbf{Step1: While} (not converged), \textbf{do }

   \textbf{Step2:} Update $\beta^{k+1}$ according to Eq.(\ref{ADMM2})

   \textbf{Step3:} Update $\gamma_{i}^{k+1},i=1,2,...,d$ according to Eq.(\ref{ADMM3})

   \textbf{Step4:} Update $\beta_{i}^{k+1},i=1,2,...,d$ according to Eq.(\ref{ADMM4})

\textbf{End While}

\textbf{Output}: $\beta^{*}$.
}
\label{algorithm2}
\end{algorithm}

\subsection{Complexity Analysis and Convergence Proof}
In this subsection, we provide an analyses of the properties of the proposed structurally interacting elastic net method. We commerce by presenting the computational complexity which is followed by a convergence analysis.
\subsubsection{Analysis of Computational Complexity}
Let $N$ be the number of features, $M$ the number of samples, and $K$ the required number of iterations to converge. At each iteration, the computational complexity for updating $\bold \beta$ according to Eq.(\ref{ADMM2}) is $O(N^{2}M)$. Additionally, the computational costs for updating $\gamma$ in Eq.(\ref{ADMM3}) and $z$ in Eq.(\ref{ADMM4}) are both $O(N)$. Therefore, the overall time complexity of the ADMM algorithm is calculated as $\max\{O(N^{2}MK),O(MK)\}$.

\subsubsection{Convergence Proof}
To theoretically prove the convergence of the ADMM algorithm, we present the following analysis.

\textbf{Theorem 1.}
Assume the iterative sequences generated by the ADMM algorithm are denoted as $\{\beta^{k}\}$, $\{\gamma^{k}\}$,and $\{z^{k}\}$, respectively. Suppose as $k$ tends to infinity, the sequence $\{z^{k}\}$ converges to a point $z^{\prime}$, that is, $\lim_{k\rightarrow\infty}z^{k}=z^{\prime}$. Following this, every limit point $(\beta^{\prime},\gamma^{\prime})$ of the iteration sequence $\{\beta^{k},\gamma^{k}\}$, together with $z^{\prime}$, satisfy the necessary first order conditions of the problem (\ref{INT1}), that is

(1)Primal feasibility, i.e., $\beta^{\prime}-\gamma^{\prime}=0$.

(2)Dual feasibility, i.e., $\nabla f(\beta^{\prime})+z^{\prime}=0$ and $0\in\partial g(\gamma^{\prime})-z^{\prime}$, where $\partial$ denotes the sub-differential operator.

We can easily prove \textbf{Theorem 1} by following a proof similar to that of Proposition 3 in Magn\'{u}sson\cite{DBLP:journals/tcns/MagnussonWRF16}. We can conveniently draw the conclusion from \textbf{Theorem 1} that, in general, the ADMM algorithm converges to a local optimum solution to problem (\ref{INT1}), that is, $(\beta^{\prime},\gamma^{\prime},z^{\prime})=(\beta^{\ast},\gamma^{\ast},z^{\ast})$.

\section{Experimental Results and Discussion}\label{s4}
In this section, we conduct a series of experiments to demonstrate the performance of the proposed structurally interacting elastic net feature selection method (InElasticNet). A comprehensive experimental study on two types of datasets is conducted to validate its effectiveness and make comparison with several state-of-the-art feature selection methods.
\subsection{Experiments on Standard Machine Learning Datasets}
Two categories of public datasets are used in our experiment, including eight machine learning (ML) datasets and two public available datasets. The ML datasets are the USPS handwritten digit data set~\cite{DBLP:journals/pami/Hull94}, Isolet speech data set and Pie data set from the UCI Machine Learning Repository~\cite{UCI}, YaleB face data set~\cite{DBLP:journals/pami/GeorghiadesBK01}, Lymphoma and Leukemia datasets~\cite{DBLP:journals/pr/VinhZCB16}, BASEHOCK and RELATHE. Note that the last two datasets are both large in feature dimension and sample size. Detailed information for these data sets are summarized in Table~\ref{Datasets}.

\begin{table*}
\vspace{-0pt}
\centering {
\tiny
\caption{Statistics of data sets used in the experiments}\label{Datasets}
\vspace{0pt}

\begin{tabular}{|c|c|c|c|}

 \hline
~Name ~      ~ &~Feature Dimension~     &~Sample Number~  &~Class Number~   \\ \hline

~USPS~      &~256~      &~9298~     &~10~   \\ \hline

~Isolet1~   &~617~      &~1560~     &~26~   \\ \hline

~Pie~       &~1024~     &~11554~    &~68~  \\ \hline

~YaleB~     &~1024~     &~2414~     &~38~  \\ \hline

~Lymphoma~  &~4026~     &~96~       &~9~   \\ \hline



~Leukemia~  &~7129~     &~73~       &~2~  \\ \hline

~BASEHOCK~  &~4862~     &~1993~     &~2~  \\ \hline

~RELATHE~   &~4322~     &~1427~     &~2~  \\ \hline

\hline
\end{tabular}
}\vspace{-0pt}
\end{table*}

To evaluate the discriminative capabilities of the information captured by our method, we compare the classification results obtained using the selected features from our proposed method with several state-of-the-art feature selection methods including Lasso~\cite{Lasso}, ULasso~\cite{DBLP:conf/aaai/ChenDLX13}, Fused Lasso~\cite{FusedLasso}, Elastic Net~\cite{ElasticNet}, Group Lasso~\cite{DBLP:journals/bmcbi/MaSH07}, InLasso~\cite{DBLP:journals/prl/ZhangTBXH17}, and one graph-based feature selection methods, namely,  GF-RW~\cite{DBLP:conf/gbrpr/CuiJB0H17}.

a) Lasso~\cite{Lasso}: As a typical sparse feature selection method, Lasso performs feature selection through the $l_{1}$-norm, where features corresponding to zero coefficients in the parameter vector will be discarded.

b) ULasso~\cite{DBLP:conf/aaai/ChenDLX13}: The uncorrelated Lasso (ULasso) aims to conduct variable de-correlation and variable selection simultaneously, so that the variables selected are uncorrelated as much as possible.

c) Fused Lasso~\cite{FusedLasso}: The fused lasso enforces sparsity in both the coefficients and their successive differences. It is desirable for applications with features ordered in some meaningful way.

d) Group Lasso~\cite{DBLP:journals/bmcbi/MaSH07}: The group Lasso is known to enforce the sparsity on variables at an inter-group level, where variables from different groups are competing to survive.

e) Elastic Net~\cite{ElasticNet}: In statistics, the elastic net is a regularized regression method that linearly combines the $l_{1}$ and $l_{2}$ penalties of the Lasso and Ridge methods. This ensures democracy among groups of correlated groups and allows selection of the relevant groups while simultaneously promoting sparse solutions for feature selection.

f) InLasso~\cite{DBLP:journals/prl/ZhangTBXH17}: Is a Lasso-type regression model which incorporates high-order feature interactions, InLasso can effectively evaluate whether a feature is redundant or interactive based on a neighborhood dependency measure. This method can avoid discarding some valuable features arising in individual feature combinations.

g) GF-RW~\cite{DBLP:conf/gbrpr/CuiJB0H17}: Is a graph-based feature selection method which incorporates pairwise relationship between samples of each feature dimension.

Generally, we adopt a 10-fold cross-validation method associated with C-SVM to evaluate the classification accuracy. To be specific, we first partition the entire sample randomly into 10 subsets (each subset with roughly equal size) and then we choose one subset for testing and use the remaining 9 subsets for training. We repeat this procedure for 10 times. The final accuracy is computed by averaging of the accuracies from all 10 experiments and we also compute the associated standard error.

\begin{figure}
\centering
\subfigure[YaleB dataset]{\includegraphics[width=0.49\linewidth]{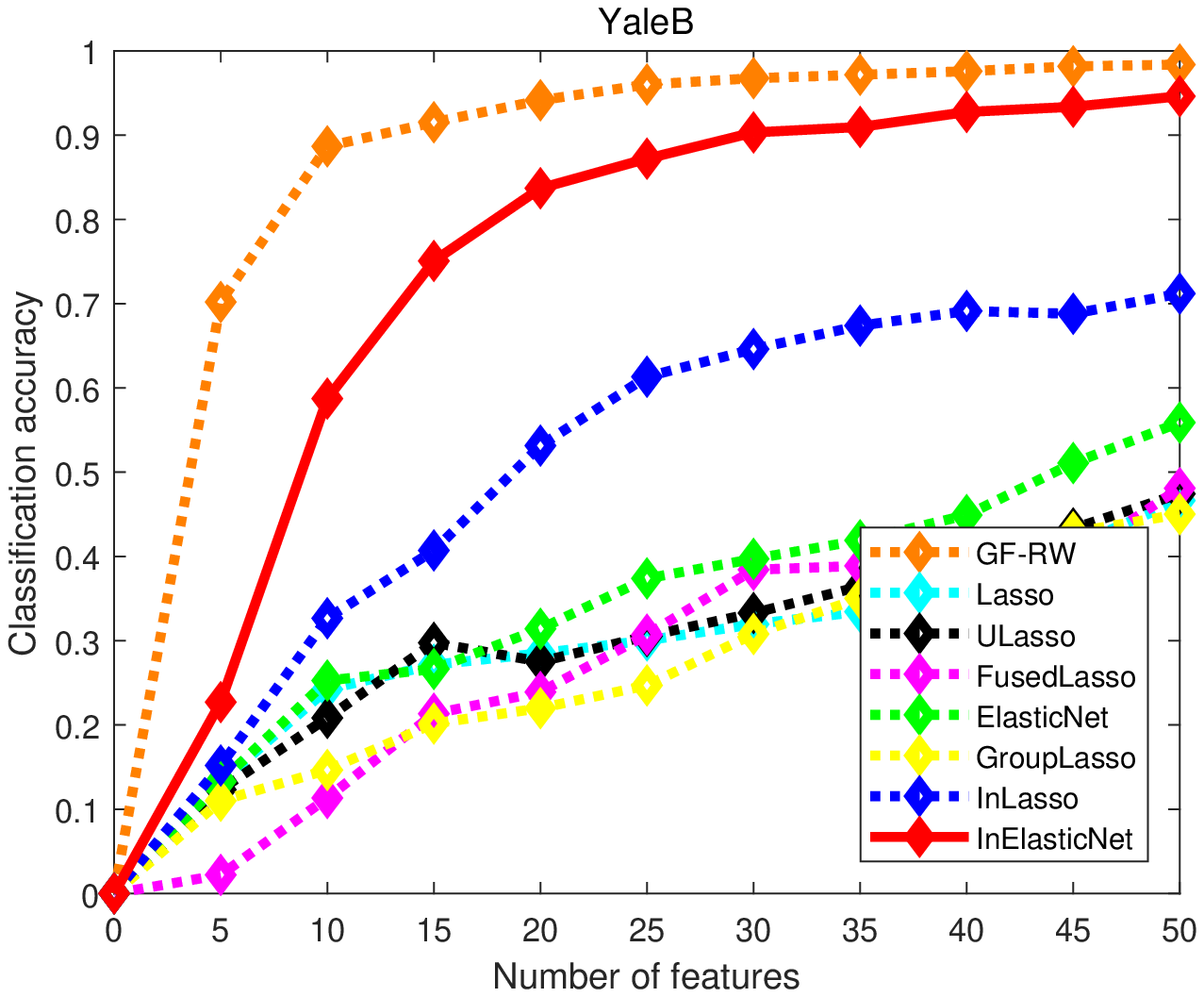}}
\subfigure[USPS dataset]{\includegraphics[width=0.49\linewidth]{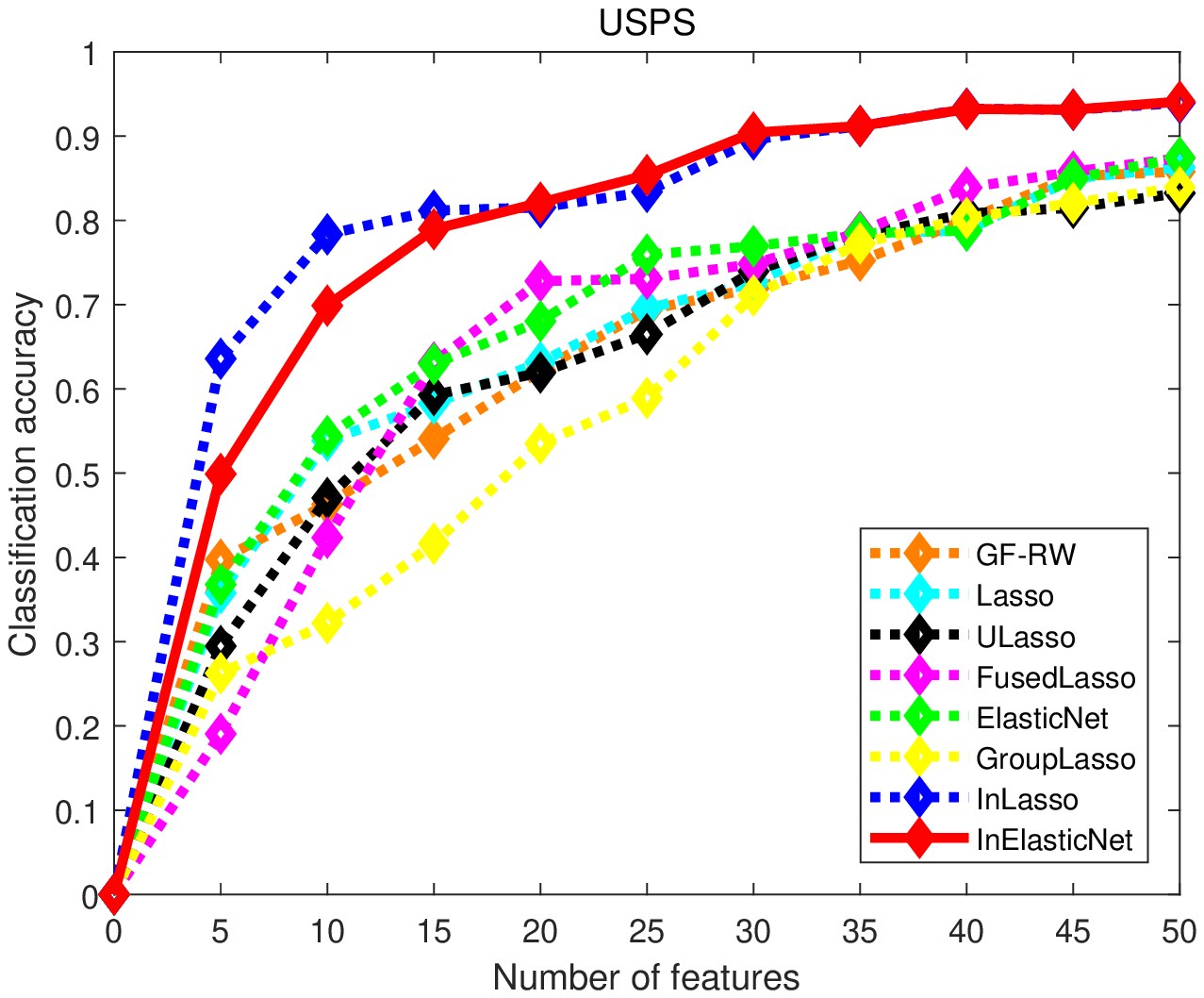}}
\subfigure[Isolet1 dataset]{\includegraphics[width=0.49\linewidth]{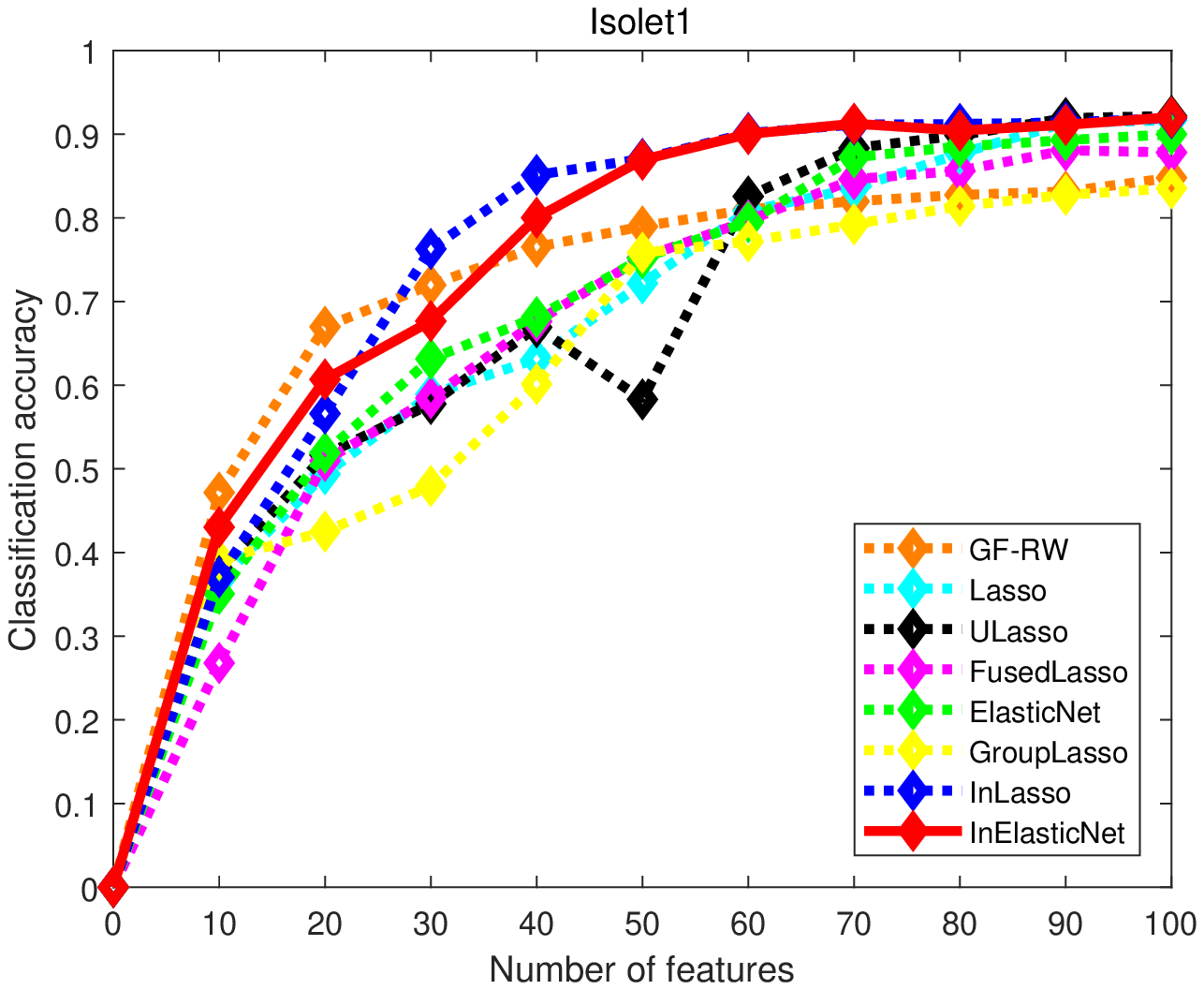}}
\subfigure[Lymphoma dataset]{\includegraphics[width=0.49\linewidth]{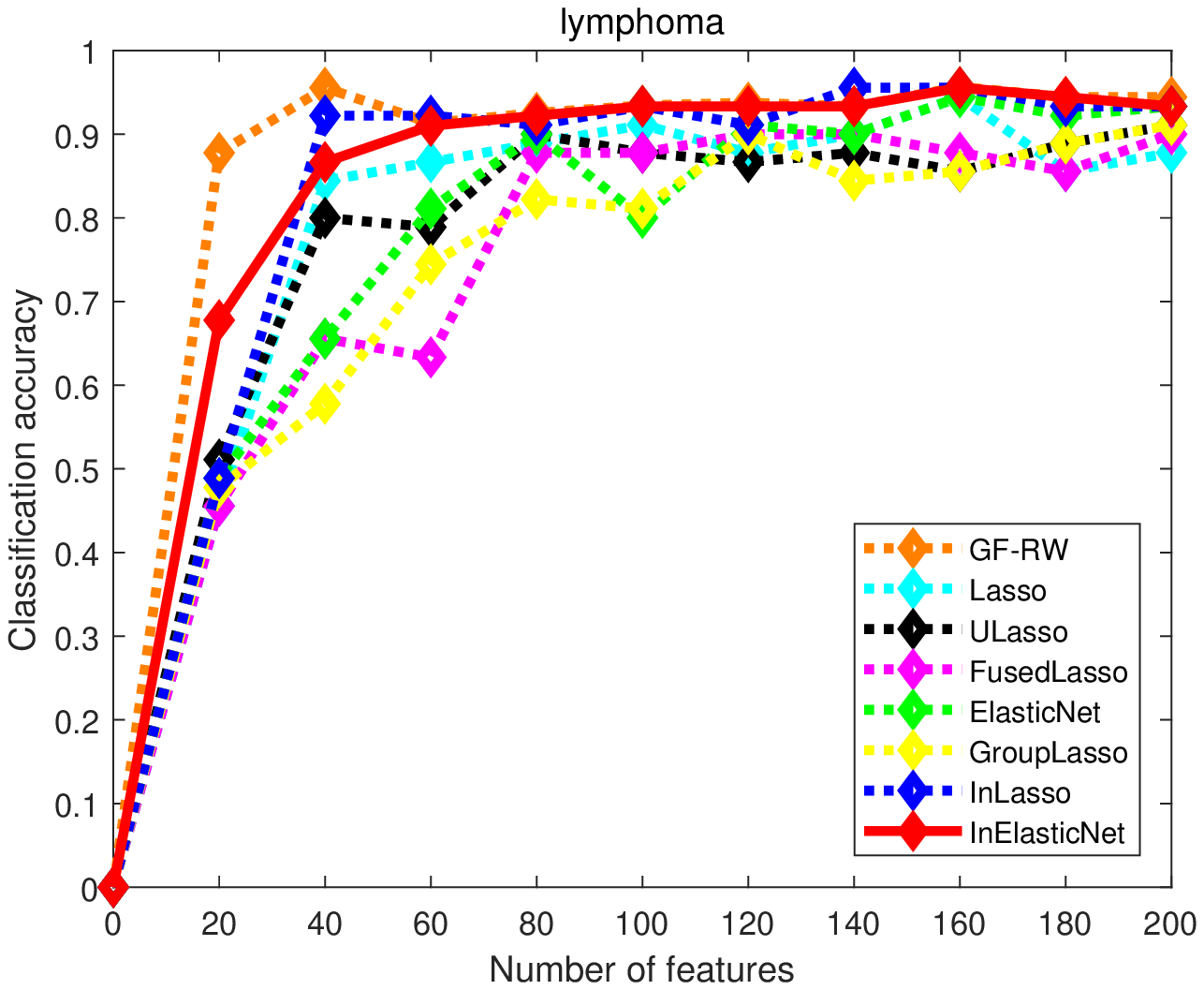}}
\subfigure[Pie dataset]{\includegraphics[width=0.49\linewidth]{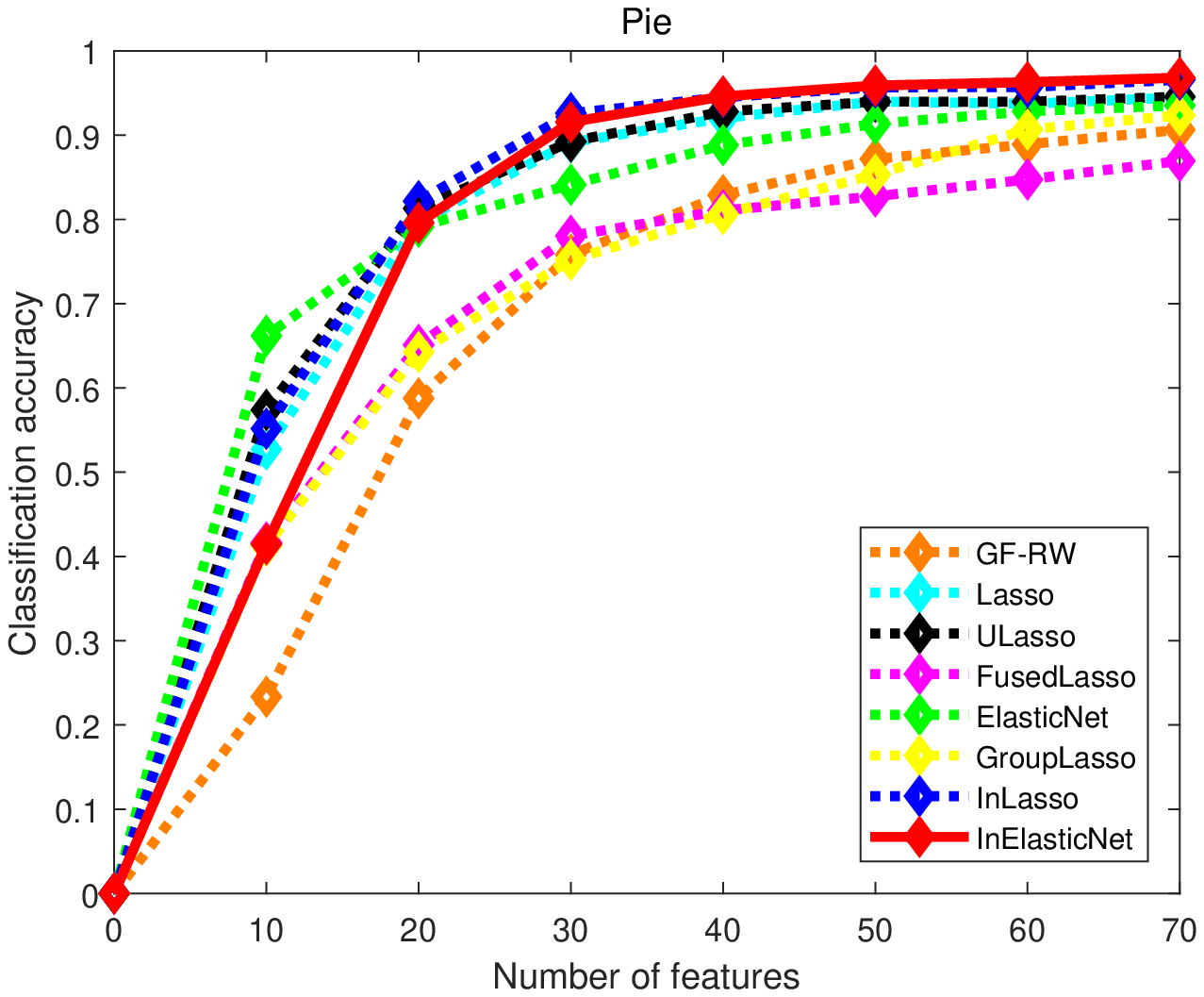}}
\subfigure[Leukemia dataset]{\includegraphics[width=0.49\linewidth]{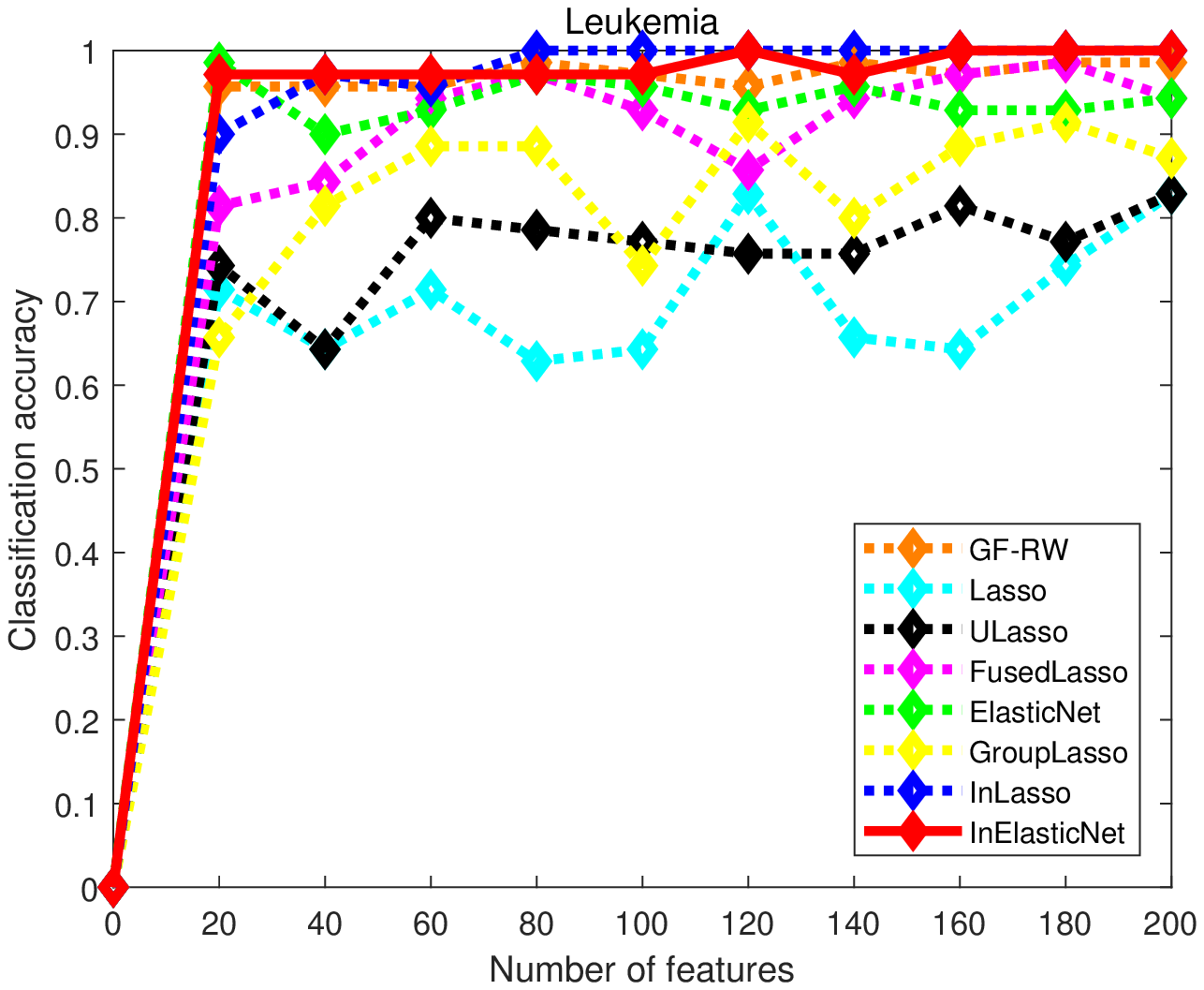}}
\subfigure[BASEHOCK dataset]{\includegraphics[width=0.49\linewidth]{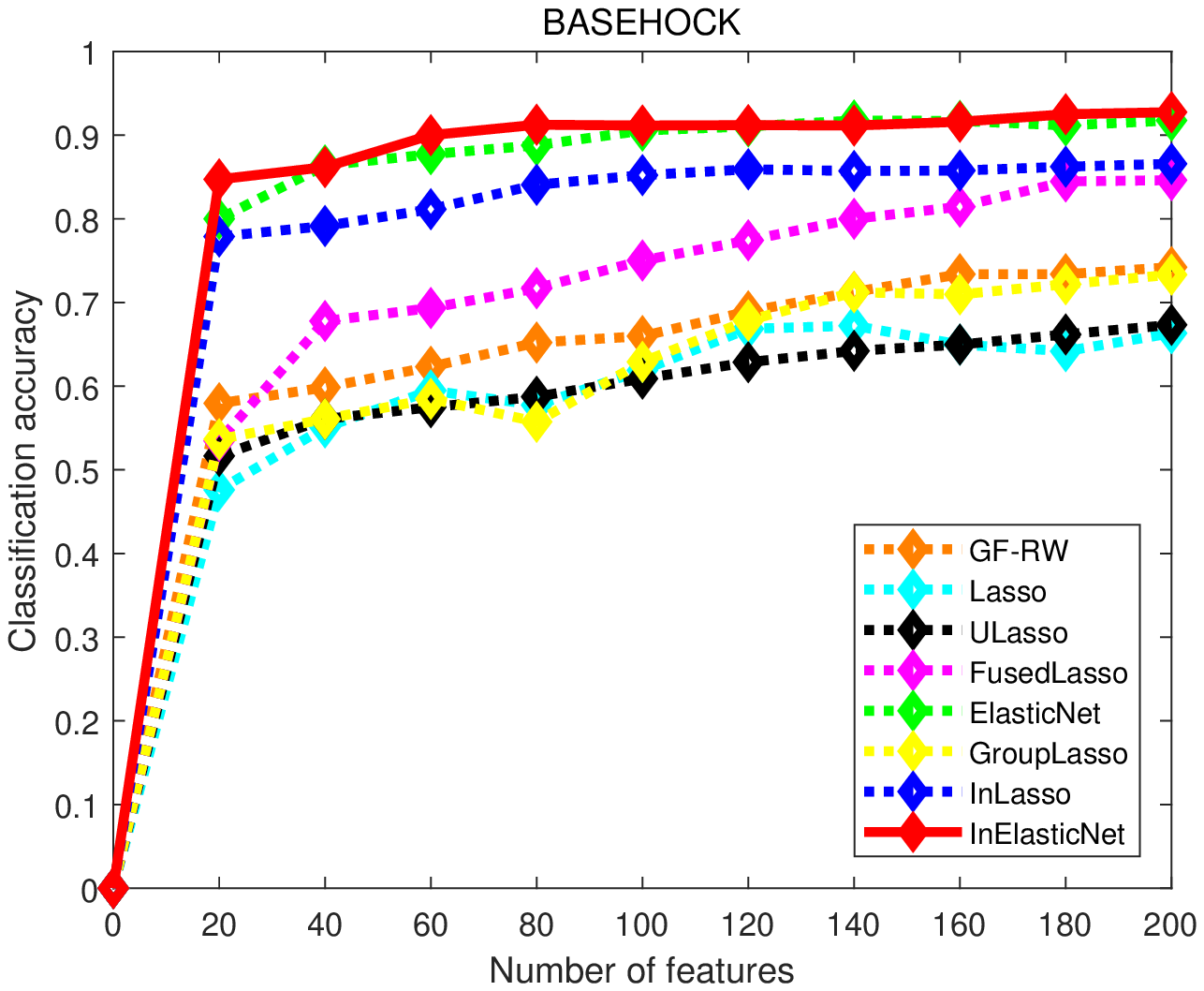}}
\subfigure[RELATHE dataset]{\includegraphics[width=0.49\linewidth]{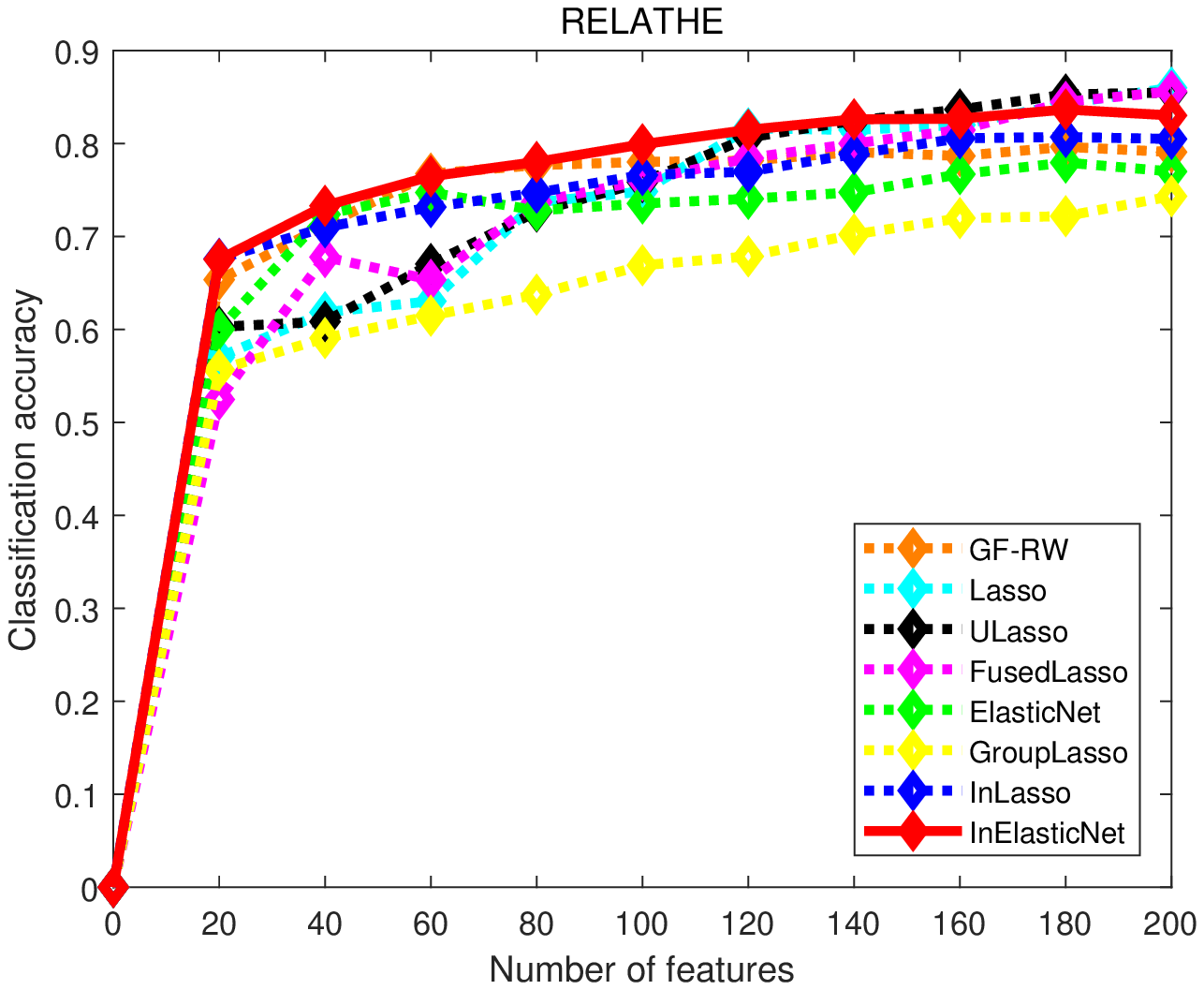}}
\vspace{-0pt}
\caption{Accuracy rate vs. the number of selected features on 8 benchmark machine learning datasets} \label{embeddingsB}
\vspace{-0pt}
\end{figure}

\begin{table*}
\vspace{-0pt}
\centering {
\tiny
\caption{Comparison of classification accuracy}\label{ComAccuracy}
\vspace{0pt}

\begin{tabular}{|c||c|c|c|c|c|c|c|c|}

  \hline
~Dataset~ &~Lasso~      &~ULasso~    &~FusedLasso~ &~ElasticNet~&~GroupLasso~&~InLasso~  &~GF-RW~ &~InElasticNet~  \\ \hline

~USPS~       &~68.17\%~    &~66.19\%~   &~68.09\%~  &~70.47\%~    &~60.71\%~  &~\textbf{84.91}\%~         &~66.99 ~ &~82.83\%~ \\
~~           &~$\pm2.61$~  &~$\pm1.61$~  &~$\pm1.40$~  &~$\pm1.83$~    &~$\pm1.28$~  &~$\pm1.05$~      &~$\pm 0.17$ ~ &~$\pm0.11$~ \\ \hline

~Isolet1~    &~71.53\%~    &~73.49\%~   &~70.49\%~  &~72.80\%~    &~66.97\%~  &~\textbf{79.79}\%~     &~75.53~  &~78.82\%~   \\
~~           &~$\pm3.70$~  &~$\pm3.29$~  &~$\pm3.19$~  &~$\pm3.93$~    &~$\pm3.23$~  &~$\pm2.95$~      &~$\pm 0.39$~  &~$\pm0.33$~ \\ \hline

~Pie~        &~85.11\%~    &~86.18\%~   &~79.77\%~  &~85.14\%~    &~75.67\%~  &~\textbf{87.48}\%~        &~75.21 ~  &~84.73\%~   \\
~~           &~$\pm0.99$~  &~$\pm0.97$~  &~$\pm1.05$~  &~$\pm0.95$~    &~$\pm0.96$~  &~$\pm0.87$~      &~$\pm 0.91$~   &~$\pm0.10$~ \\ \hline

~YaleB~      &~31.62\%~    &~34.93\%~   &~29.52\%~  &~36.75\%~    &~28.44\%~  &~54.42\%~                 &~\textbf{92.86}~   &~78.94\%~  \\
~~           &~$\pm2.77$~  &~$\pm2.74$~  &~$\pm2.52$~  &~$\pm3.90$~    &~$\pm2.71$~  &~$\pm2.67$~      &~$\pm 0.49$~   &~$\pm0.25$~ \\ \hline

~Leukemia~   &~70.43\%~    &~76.71\%~   &~92.00\%~  &~94.29\%~    &~83.71\%~  &~98.29\%~     &~97.14~  &~\textbf{98.29}\%~   \\
~~           &~$\pm1.75$~  &~$\pm1.63$~  &~$\pm1.26$~  &~$\pm1.05$~    &~$\pm1.53$~  &~$\pm0.33$~      &~$\pm 1.67$ ~ &~$\pm1.73$~ \\ \hline

~Lymphoma~   &~84.22\%~    &~82.78\%~   &~79.34\%~  &~82.67\%~    &~78.33\%~  &~88.67\%~               &~\textbf{93.23}~  &~90.10\%~   \\
~~           &~$\pm1.19$~  &~$\pm1.30$~  &~$\pm1.40$~  &~$\pm1.31$~    &~$\pm1.49$~  &~$\pm0.95$~      &~$\pm 0.97$~  &~$\pm1.03$~ \\  \hline

~BASEHOCK~   &~61.23\%~    &~61.04\%~   &~74.53\%~  &~89.09\%~    &~66.34\%~  &~83.77\%~               &~67.25~  &~\textbf{90.25}\%~   \\
~~           &~$\pm0.37$~  &~$\pm0.39$~  &~$\pm0.28$~  &~$\pm0.35$~    &~$\pm0.35$~  &~$\pm0.33$~      &~$\pm 0.38$~   &~$\pm0.32$~ \\  \hline

~RELATHE~    &~74.58\%~    &~75.42\%~   &~74.57\%~  &~73.39\%~    &~76.07\%~  &76.04\%~                                         &~76.36 ~  &~\textbf{78.87}\%~   \\
~~           &~$\pm0.35$~  &~$\pm0.49$~  &~$\pm0.44$~  &~$\pm0.42$~    &~$\pm0.41$~  &~$\pm0.24$~      &~$\pm 0.40$~   &~$\pm0.23$~ \\ \hline

~AVG~        &~68.35\%~    &~69.60\%~   &~71.04\%~  &~75.57\%~    &~67.03\%~  &~81.62\%~                    &~80.54\% ~  &~\textbf{85.36}\%~   \\

\hline
\end{tabular}
}\vspace{-0pt}
\end{table*}

Fig.~\ref{embeddingsB} shows the classification accuracy versus the number of selected features on the datasets for different methods. It is clear from the figure that the proposed method InElasticNet is, by and large, superior to the following alternative feature selection methods including Lasso, ULasso, Elastic Net, Fused Lasso, and Group Lasso on all datasets. When compared with InLasso, it is clear that our method significantly outperforms InLasso on the YaleB dataset and also outperforms InLasso for BASEHOCK and RELATHE, which are challenging datasets both large in feature dimension and sample size. For the remaining datasets, our method is competitive to InLasso. In addition, our method significantly outperforms GF-RW method on USPS, Pie and BASEHOCK, and is superior than GF-RW on Isolet1, Leukemia, and RELATHE datasets. As Fig.~\ref{embeddingsB} shows, when the number of selected features is too small, the advantage of our method is not clear. However, when the number of total features in the selected subset increases to a certain number, the InElasticNet method performs much better than the alternative methods. The results verify that the proposed structurally interacting Elastic Net can identify more informative feature subsets than the state-of-the-art feature selection methods. Although the total numbers of features in the selected feature subsets are on average slightly larger than those obtained via Lasso, it is still small as compared to the total number of features in the datasets. This is because our method can select correlated features and also encourage a sparse solution, whereas Lasso tends to select only one feature from a group of correlated features, which may decrease its accuracy.

To make a detailed comparison, we report the mean classification accuracies and corresponding variances (i.e.,MEAN$\pm$ STD) obtained via various methods on each data set with different number of features selected by using the C-SVM classifier in Table~\ref{ComAccuracy}. The mean classification accuracy is obtained by averaging the accuracy achieved via C-SVM using the largest number of features indicated in the corresponding subfigures of Fig.~\ref{embeddingsB} for each data set. For instance, for the YaleB dataset, we use the top 10, 20,...,50 features selected by each algorithm. The boldfaced value of each row corresponds to the highest accuracy obtained by the different methods for the underlying dataset. Our proposed method, i.e., InElasticNet improves the classification accuracy by 24.52\% for the YaleB dataset and 1.43\% for the Lymphoma dataset, respectively. For the Leukemia dataset, InElasticNet performs equally to InLasso, and is better than the alternative methods. As for Isolet1, USPS, and Pie, InElasticNet can obtain better classification accuracy than Lasso, ULasso, FusedLasso, Elastic Net, and Group Lasso, and is competitive to InLasso, which retains the highest classification accuracy. Additionally, for the two large datasets Basehock and Relathe, InElasticNet outperforms all the competitors.

The bottom row of Table~\ref{ComAccuracy} displays the average classification accuracy for each algorithm over the eight datasets. It shows that our proposed method, i.e., InElasticNet improves the classification accuracy by 17.11\%(Lasso), 15.57\%(ULasso), 15.75\%(FusedLasso), 11.93\%(ElasticNet), 19.98\%(GroupLasso), 3.36\%(InLasso), and 4.82\%(GF-RW), respectively, compared to the averaged classification accuracy of all alternative methods on the eight datasets. In addition, it is worth noting that the standard errors for the proposed InElasticNet method are smaller than the competing methods for almost all datasets, except for Leukemia. This indicates that InElasticNet is more stable than the competing methods.

%
%

\begin{table*}[t!]
\centering {
\tiny
\caption{Win/Tie/Lost matrix for the feature selection methods used in the experiments.}\label{WinLoss}
\begin{tabular}{|c||c|c|c|c|c|c|c|c||r|}
  \hline
  ~ Methods      ~&~Lasso    ~&~ ULasso ~&~FusedLasso~&~ElasticNet~&~GroupLasso~&~InLasso  ~&~GF-RW    ~&~InElasticNet ~&~ Total ~ \\ \hline
  \hline
  ~ Lasso        ~&~         ~&~ $0/6/2$ ~&~ $3/3/2$ ~&~ $0/4/4$  ~&~ $5/1/2$  ~&~ $0/1/7$ ~&~ $1/2/5$ ~&~ $0/1/7$     ~&~ $~~9/18/29$ ~ \\ \hline
  ~ ULasso       ~&~ $2/6/0$ ~&~         ~&~ $4/2/2$ ~&~ $1/4/3$  ~&~ $5/1/2$  ~&~ $0/2/6$ ~&~ $1/2/5$ ~&~ $0/1/7$     ~&~ $13/18/25$ ~ \\ \hline
  ~ FusedLasso   ~&~ $2/3/2$ ~&~ $2/2/4$ ~&~         ~&~ $0/1/7$  ~&~ $5/3/0$  ~&~ $0/1/7$ ~&~ $2/2/4$ ~&~ $0/0/8$     ~&~ $11/12/33$ ~ \\ \hline
  ~ ElasticNet   ~&~ $4/4/0$ ~&~ $3/4/1$ ~&~ $7/1/0$ ~&~          ~&~ $7/1/0$  ~&~ $1/1/6$ ~&~ $3/0/5$ ~&~ $0/2/6$     ~&~ $25/13/18$ ~ \\ \hline
  ~ GroupLasso   ~&~ $2/1/5$ ~&~ $2/1/5$ ~&~ $0/3/5$ ~&~ $0/1/7$  ~&~          ~&~ $0/1/7$ ~&~ $0/3/5$ ~&~ $0/0/8$     ~&~ $~~4/10/42$ ~ \\ \hline
  ~ InLasso      ~&~ $7/1/0$ ~&~ $6/2/0$ ~&~ $7/1/0$ ~&~ $6/1/1$  ~&~ $7/1/0$  ~&~         ~&~ $4/2/2$ ~&~ $2/3/3$     ~&~ $39/11/~~6$ ~ \\ \hline
  ~ GF-RW        ~&~ $5/2/1$ ~&~ $5/2/1$ ~&~ $4/2/2$ ~&~ $5/0/3$  ~&~ $5/3/0$  ~&~ $2/2/4$ ~&~         ~&~ $2/1/5$     ~&~ $28/12/16$ ~ \\ \hline
  ~ InElasticNet ~&~ $7/1/0$ ~&~ $7/1/0$ ~&~ $8/0/0$ ~&~ $6/2/0$  ~&~ $8/0/0$  ~&~ $3/3/2$ ~&~ $5/1/2$ ~&~             ~&~ $\textbf{44/~~8/~~4}$ ~ \\ \hline

\end{tabular}
}\vspace{-0pt}
\end{table*}

\begin{table*}
\vspace{-0pt}
\centering {
\tiny
\caption{The best result of all methods and the corresponding size of selected feature subset}\label{Bestresults}
\vspace{0pt}

\begin{tabular}{|c||c|c|c|c|c|c|c|c|}

\hline
~Dataset~      ~ &~USPS~     &~Isolet1~  &~Pie~ &~YaleB~&~Leukemia~&~Lymphoma~                                                          &~BASEHOCK~ &~RELATHE~ \\ \hline \hline

~Lasso~    &~86.30\%$(50)$~&~91.67\%$(100)$~&~94.48\%$(70)$~ &~46.64\%$(50)$~ &~82.86\%$(120)$~&~94.44\%$(160)$~  &~66.33\%$(200)$~  &~86.00\%$(200)$~\\ \hline

~ULasso~     &~83.24\%$(50)$~    &~92.18\%$(100)$~  &~94.57\%$(70)$~  &~47.43\%$(50)$~    &~82.86\%$(200)$~  &~91.11\%$(200)$~      &~67.30\%$(200)$~  &~84.62\%$(200)$~   \\ \hline

~FusedLasso~  &~87.40\%$(50)$~    &~88.08\%$(90)$~  &~93.53\%$(70)$~  &~55.89\%$(50)$~    &~98.57\%$(20)$~  &~94.44\%$(160)$~       &~84.62\%$(200)$~  &~\textbf{86.50}\%$(200)$~    \\ \hline

~ElasticNet~   &~87.43\%$(50)$~    &~90.00\%$(100)$~  &~86.94\%$(70)$~  &~48.09\%$(50)$~  &~98.57\%$(180)$~  &~90.00\%$(120)$~      &~91.76\%$(200)$~  &~77.95\%$(180)$~  \\ \hline

~GroupLasso~  &~83.93\%$(50)$~    &~83.53\%$(100)$~  &~92.35\%$(70)$~  &~45.02\%$(50)$~  &~91.43\%$(180)$~  &~91.11\%$(200)$~      &~73.33\%$(200)$~  &~74.16\%$(200)$~ \\ \hline

~InLasso~   &~93.94\%$(50)$~    &~91.92\%$(100)$~  &~96.58\%$(70)$~  &~71.20\%$(50)$~   &~\textbf{100\%}$(80)$~  &~\textbf{95.56\%}$(140)$~   &~86.58\%$(200)$~  &~80.70\%$(180)$~   \\ \hline

~GF-RW~   &~85.79\%$(50)$~    &~84.80\%$(100)$~  &~90.64\%$(70)$~  &~\textbf{98.38}\%$(50)$~    &~98.57\%$(20)$~  &~\textbf{95.56\%}$(160)$~     &~74.22\%$(200)$~  &~76.36\%$(180)$~  \\ \hline

~InElasticNet~&~\textbf{94.10\%}$(50)$~    &~\textbf{92.23\%}$(100)$~  &~\textbf{96.81}\%$(70)$~  &~94.62\%$(50)$~    &~\textbf{100\%}$(120)$~  &~\textbf{95.56\%}$(160)$~  &~\textbf{92.75}\%$(200)$~  &~83.66\%$(180)$~     \\ \hline

\end{tabular}
}\vspace{-0pt}
\end{table*}

Table~\ref{WinLoss} presents the Win/Tie/Lost matrix for the feature selection methods used in the experiments. The $(i,j)$th element of the matrix represents the number of datasets where the method corresponding to the $i$th row has won/tied/lost against the method corresponding to the $j$th column. A tie is defined as a dataset on which difference in classification accuracy between two methods is not statistically significant. The last column of Table~\ref{WinLoss} shows the total number of wins/ties/lost for a given method, and the best performing method is highlighted in bold. InElasticNet has the largest total number of wins and the smallest total number of lost. This clearly indicates that the proposed InElasticNet method performs significantly better than the alternative feature selection methods.

Table~\ref{Bestresults} shows the best results for each competing method together with their corresponding number of features in the selected subset. In the table, the best classification accuracy is shown which is followed by the optimal number of features selected in brackets. From this table, it is clear that the proposed method achieves the highest classification accuracy using same number of features in the selected subset as the alternative methods. This implies that our proposed method tremendously outperforms all the competing methods and has more discriminative power.

\begin{figure}
\centering
\subfigure[YaleB dataset]{\includegraphics[width=0.49\linewidth]{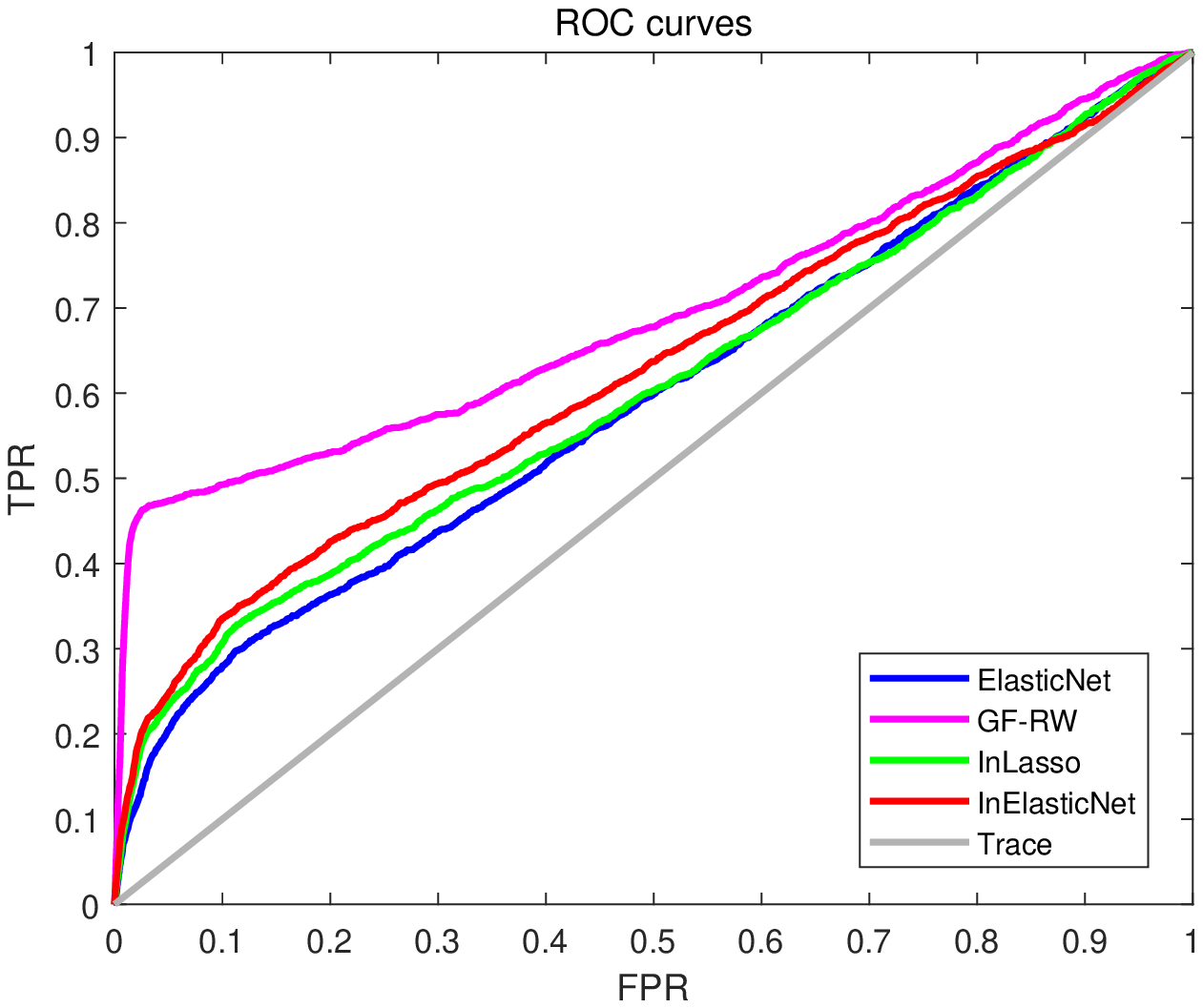}}
\subfigure[USPS dataset]{\includegraphics[width=0.49\linewidth]{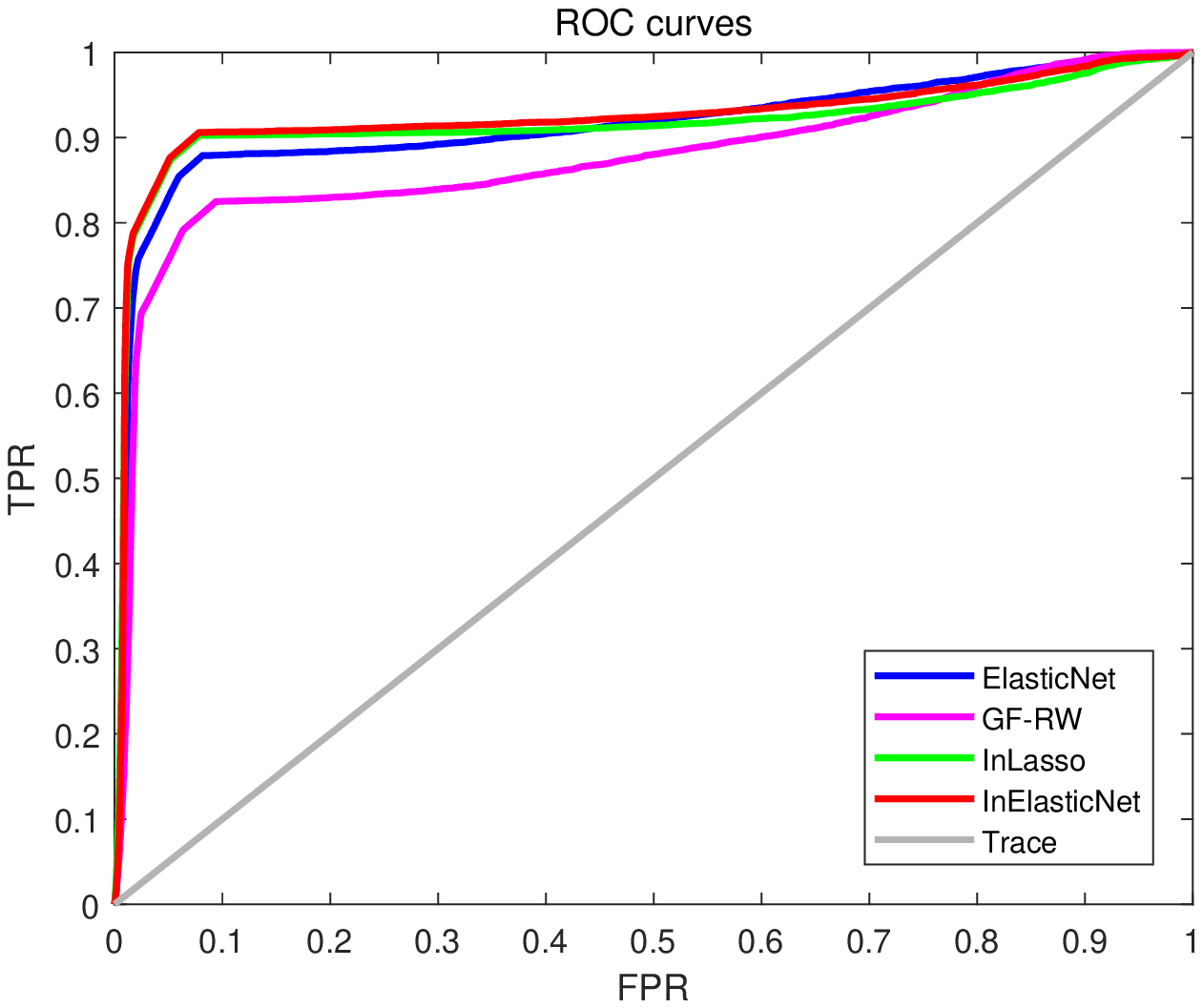}}
\subfigure[Isolet1 dataset]{\includegraphics[width=0.49\linewidth]{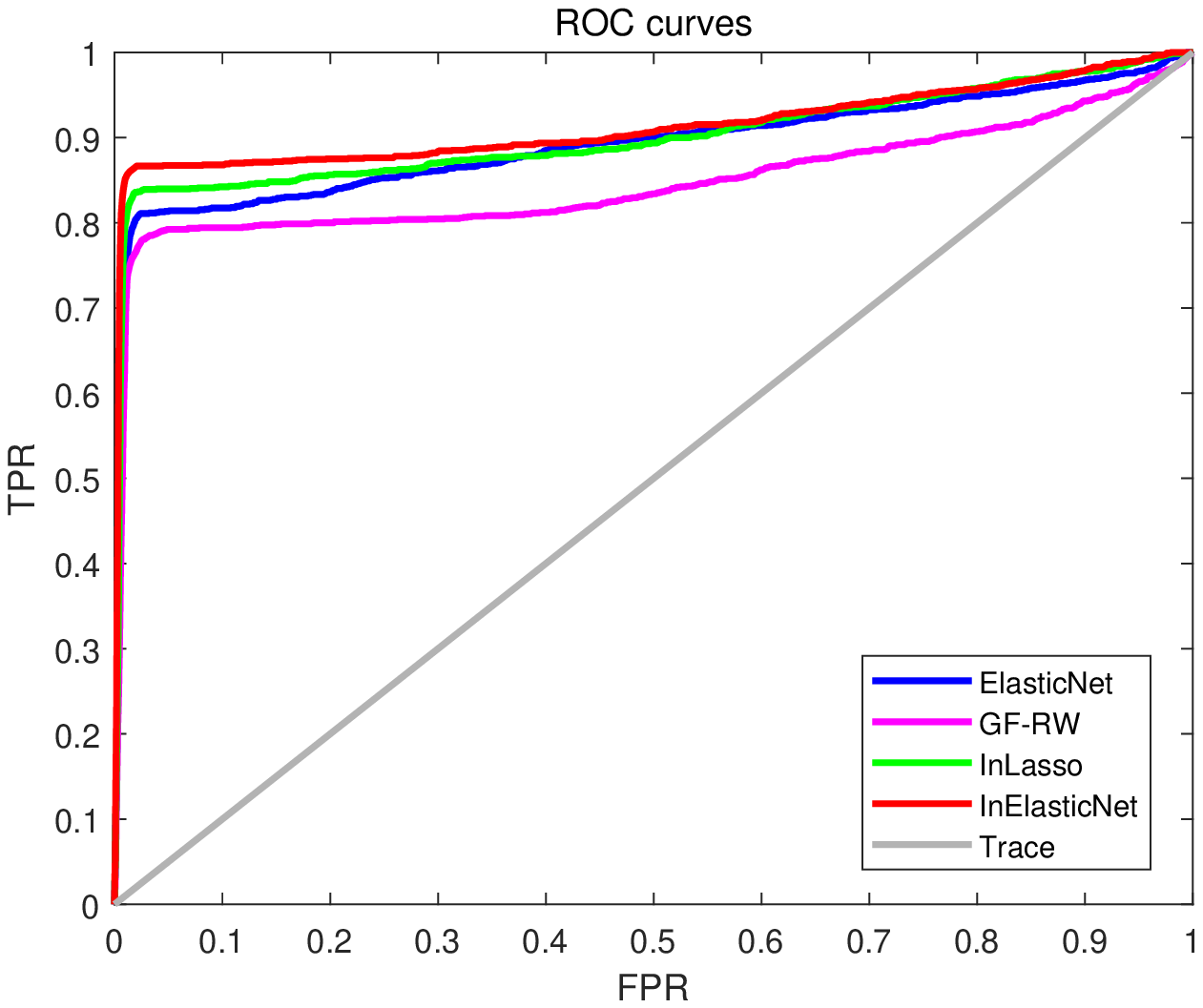}}
\subfigure[Lymphoma dataset]{\includegraphics[width=0.49\linewidth]{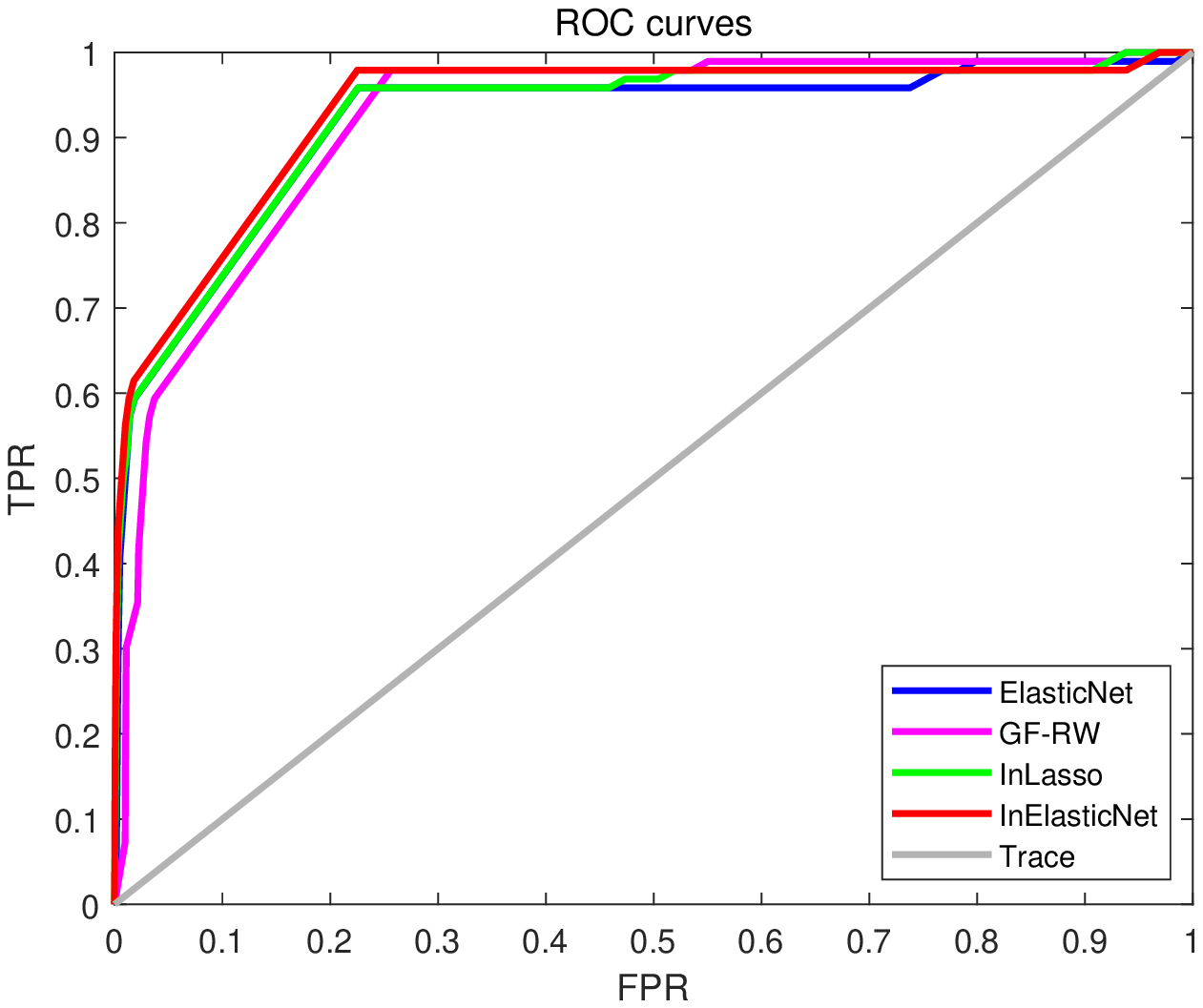}}
\subfigure[Pie dataset]{\includegraphics[width=0.49\linewidth]{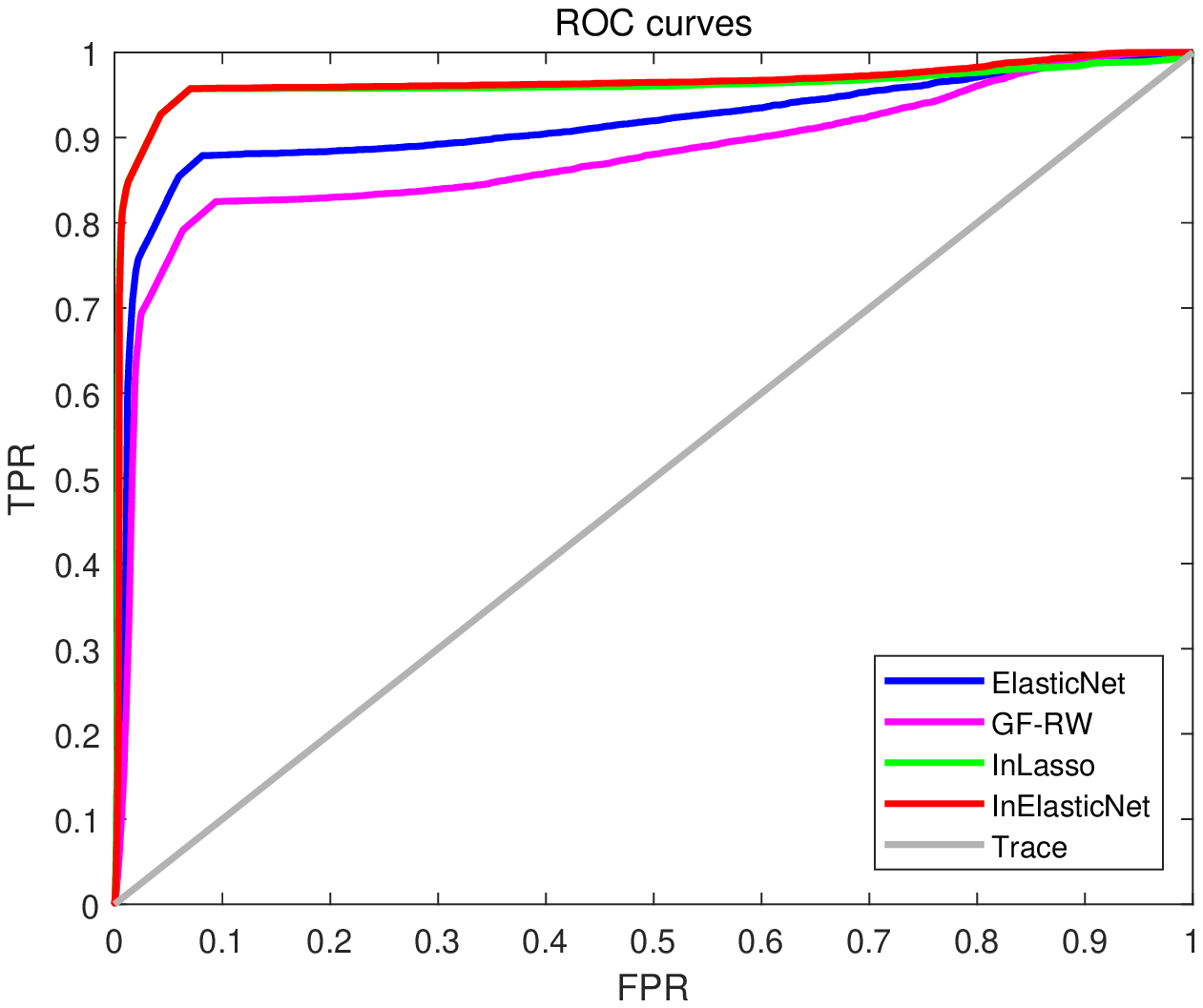}}
\subfigure[Leukemia dataset]{\includegraphics[width=0.49\linewidth]{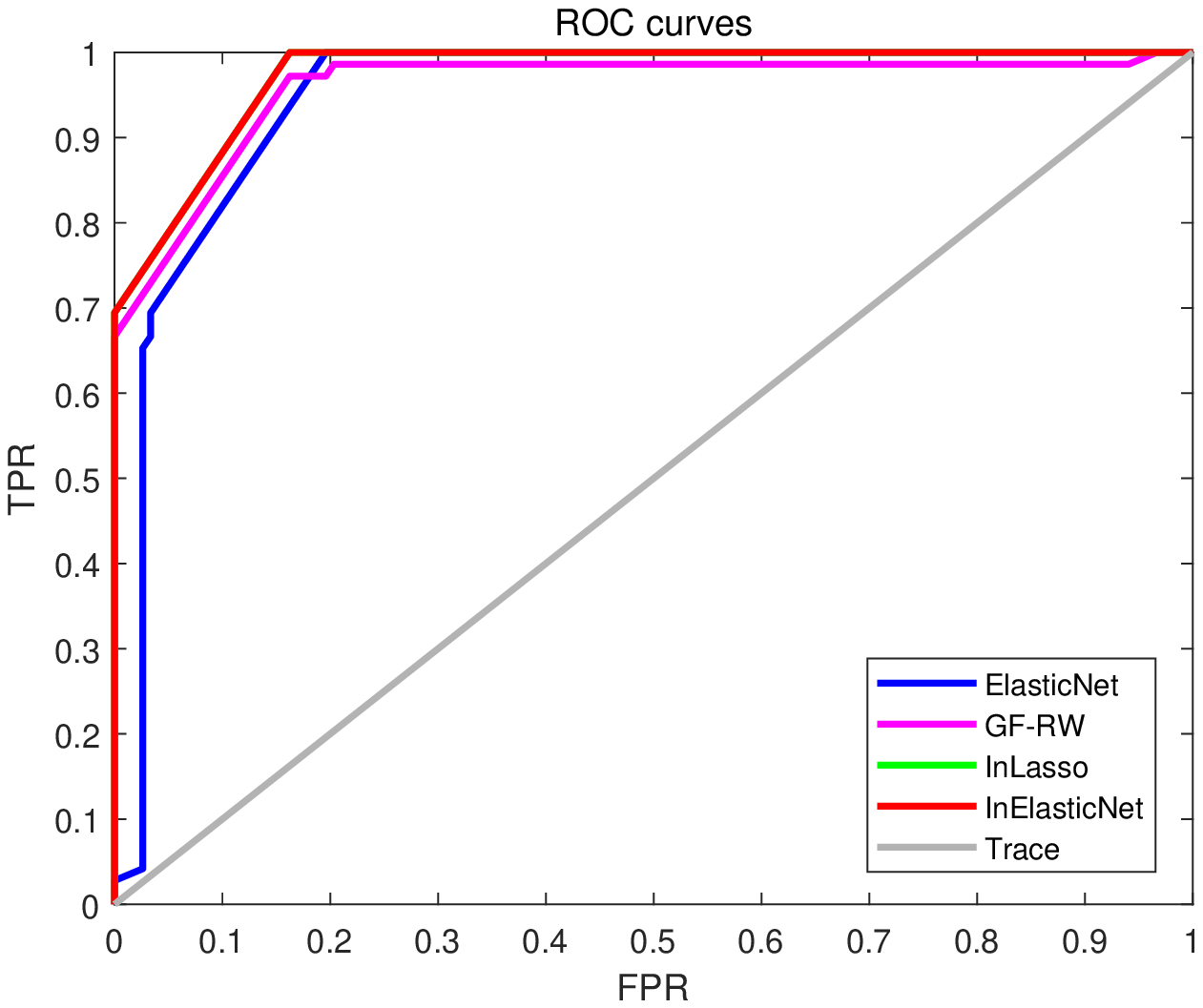}}
\subfigure[Basehock dataset]{\includegraphics[width=0.49\linewidth]{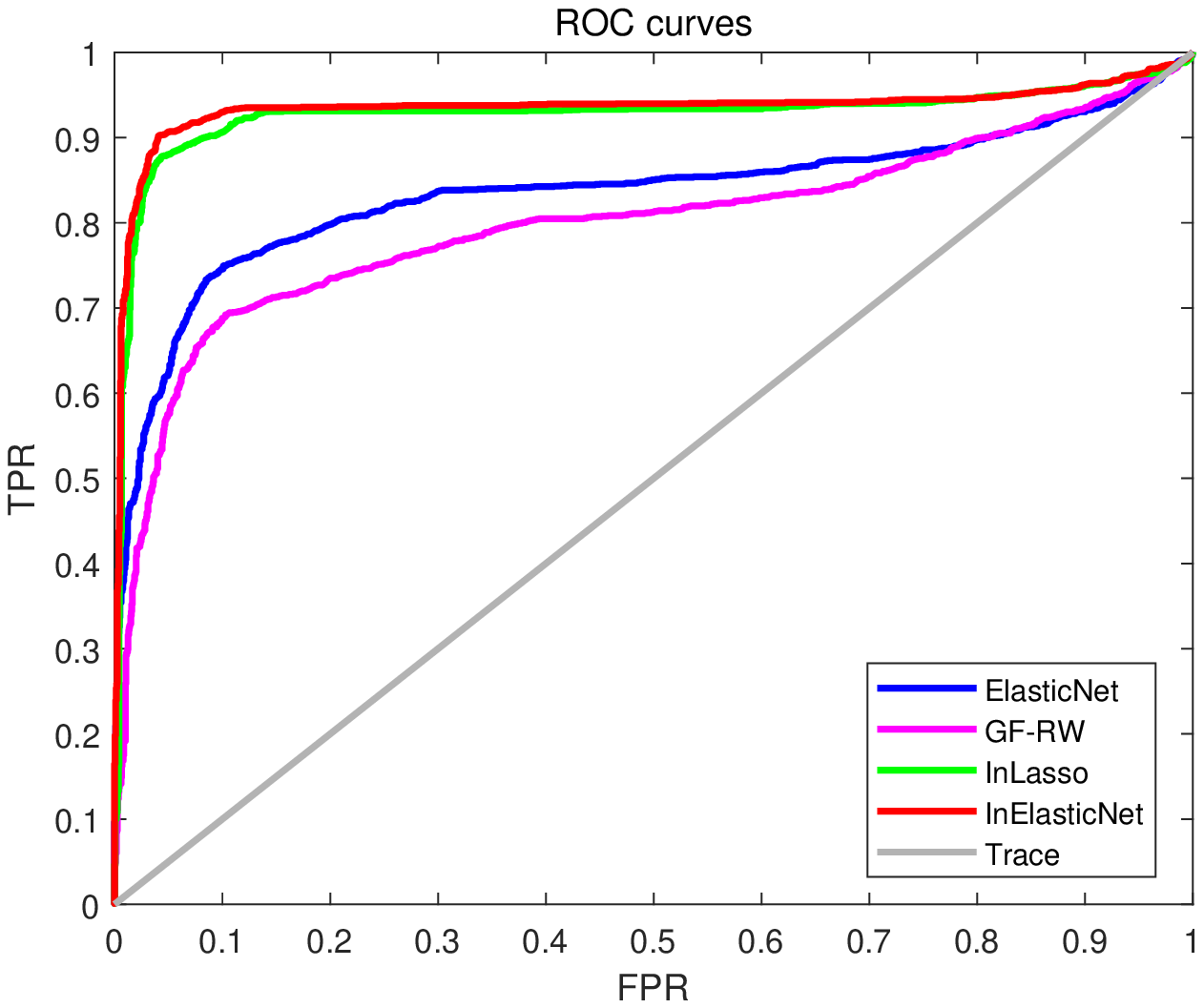}}
\subfigure[Realthe dataset]{\includegraphics[width=0.49\linewidth]{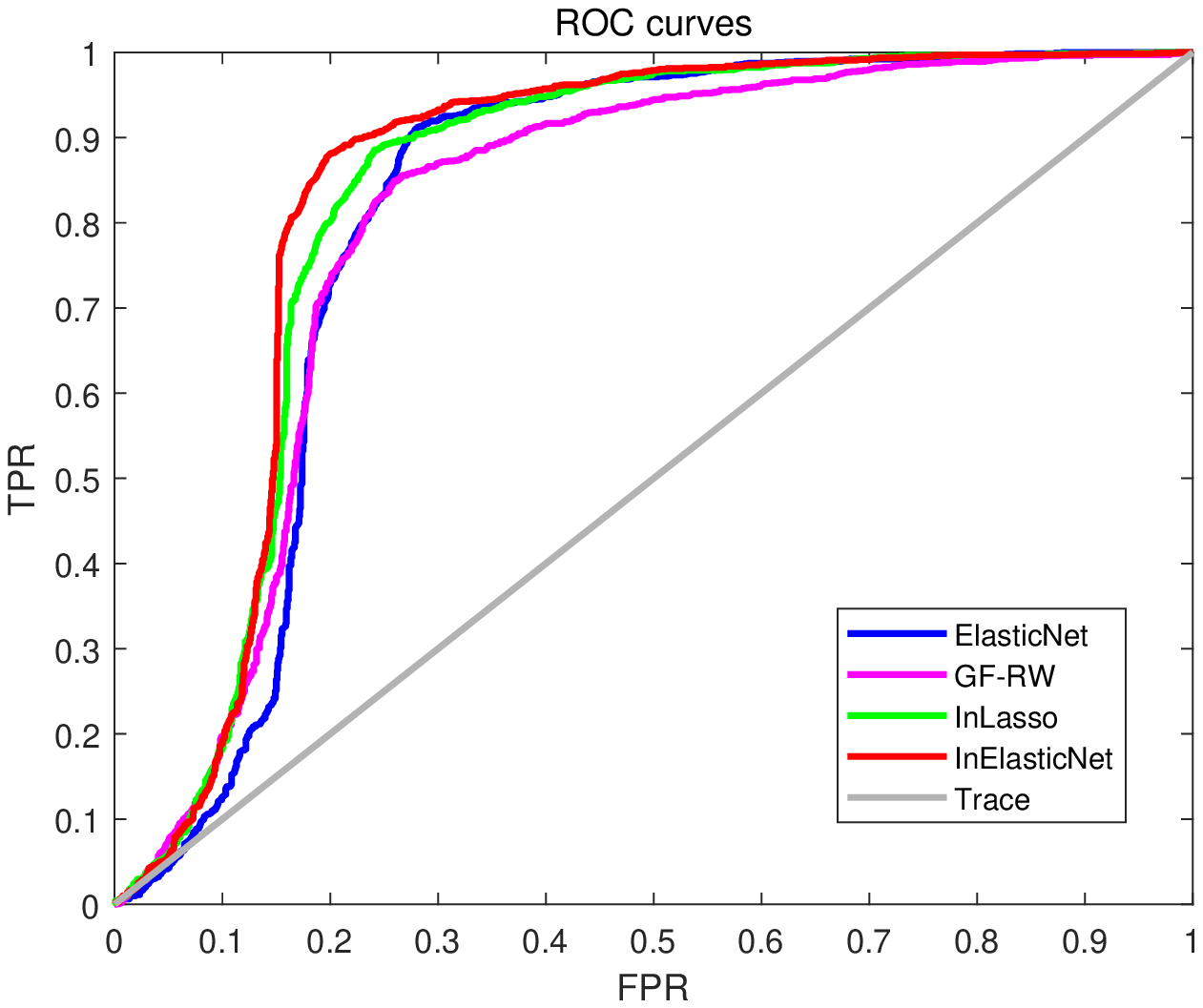}}
\vspace{-0pt}
\caption{ROC Curves for Different Datasets.} \label{roc}
\vspace{-0pt}
\end{figure}

The ROC curves of the most competitive methods (ElasticNet, GF-RW, InLasso and InElasticNet) on the eight datasets are plotted in Figure~\ref{roc}. From this figure, we observe that except for the YaleB dataset, the proposed InElasticNet method achieves superior performance to the competitors on all datasets. In summary, the aforementioned experimental results demonstrate that our proposed feature selection method outperforms the alternative methods on the standard ML datasets.

\subsection{Experiments on Two Real World Datasets}
Apart from ML datasets, two publically available datasets including i) a Peer-to-Peer (P2P) dataset collected from the P2P lending sector in China and ii) a healthcare dataset collected from a well-known medical platform is used to validate the effectiveness of the proposed feature selection approach.

Since the launch of the first P2P lending platform in 2007, the P2P lending industry has developed rapidly and the market is enormous. Specifically, the total number of operational P2P lending platforms nationwide has reached 2,448 with an accumulative loan amount of 20 trillion yuan by the end of year 2016. Along with this rapid development, the P2P lending industry has also experienced some serious problems with rising defaults and weak risk control. Therefore, it is of great significance to develop an effective decision aid for the credit risk analysis of P2P platforms. However, the P2P lending data are often high-dimensional, highly correlated and unstable. Therefore, it presents a challenge for traditional statistical pattern recognition and machine learning techniques. The aim is to effectively analyze the data and identify which factors influence the performance of the lending platforms, or the default probability of the borrowers, etc. To realize these goals, the sample relationships of the P2P data that encapsulates significant information should be incorporated into the feature selection process. However, the majority of existing feature selection methods ignore the sample relationships and may cause significant information loss. By contrast, our proposed structurally interacting feature selection approach is able to encapsulate the sample relationships of P2P data and overcome these shortcomings.

The P2P dataset is collected from a reputable P2P lending portal in China\footnote{See the website http://www.wdzj.com/ for more details}, which tracks the industry. The dataset consists of the most popular 200 platforms (i.e., 200 samples) until Aug 2014. For each platform, we collect 19 features including 1) transaction volume, 2) total turnover, 3) average annualized interest rate, 4) the total number of borrowers, 5) the total number of investors, 6) the online time, which refers to the foundation year of the platform, 7) the operation time, i.e., number of months since the foundation of the platform, 8) registered capital, 9) weighted turnover, 10) average term of loan, 11) average full mark time, i.e., tender period of a loan raised to the required full capital, 12) average amount borrowed, i.e., average loan amount of each successful borrower, 13) average amount invested, which is the average investment amount of each successful investor, 14) loan dispersion, i.e., the ratio of the repayment amount to the total capital, 15) investment dispersion, the ratio of the invested amount to the total capital, 16) average times of borrowing, 17) average times of investment, 18) loan balance, and 19) popularity.

We evaluate the performance of the proposed feature selection approach with respect to continuous target features. Specifically, we use the proposed method to perform credit risk evaluation of the P2P lending platforms. As it is difficult to obtain sufficient data for the platforms which encountered a problem, we use the annualized average interest rate as an indicator of the credit risk of the P2P lending platforms. In finance, interest rate is the amount charged, expressed as a percentage of principal, by a lender to a borrower for the use of assets. When the borrower is a low-risk party, they will usually be charged a low interest rate. On the other hand, if the borrower is considered high risk, the interest rate charged will be higher. Likewise, a higher annualized average interest rate of the P2P lending platforms often indicates a greater likelihood of default, i.e., higher credit risk of the platforms. Identifying the features most relevant to the interest rate can help investors effectively manage the credit risks involved in P2P lending. Therefore, in our experiment, we set the average annualized interest rate as the target feature which takes continuous values. Our aim is to identify the most informative subset of features for the credit risk of the P2P platforms by using the proposed feature selection method. To further strengthen our findings, we also compare the proposed feature selection method with two alternative methods including Elastic Net~\cite{ElasticNet} and Interacted Lasso~\cite{DBLP:journals/prl/ZhangTBXH17}.

\begin{table*}
\vspace{-0pt}
\centering {
\tiny
\caption{Comparison of three methods for the P2P dataset}\label{T:GraphInformation22}
\vspace{0pt}

\begin{tabular}{|c||c||c||c|}

  \hline
 ~Ranking ~      ~ &~ElasticNet~                                &~InLasso~                            & ~InElasticNet~  \\ \hline \hline
 ~1\#~    &~ $\textrm{Loan~dispersion}$  ~  &~$\textrm{Average~amount~invested}$~          &~$\textrm{Average~amount~invested}$~      \\ \hline

 ~2\#~    &~  $\textrm{Investment~dispersion}$ ~      &~$\textrm{Average~times~of~investment}$~              &~$\textrm{Average~times~of~investment}$~     \\ \hline

 ~3\#~    &~ $\textrm{Popularity}$ ~  &~$\textrm{Online~time}$~     &~$\textrm{Online~time}$~     \\ \hline

 ~4\#~    &~ $\textrm{Operation~time}$  ~            &~$\textrm{Total~number~of~investors}$~             &~$\textrm{Investment~dispersion}$~      \\ \hline

 ~5\#~    &~ $\textrm{Average~times~of~borrowing}$ ~    &~$\textrm{Average~term~of~loan}$~ &~$\textrm{Loan~balance}$~     \\ \hline

 ~6\#~    &~ $\textrm{Online~time}$~  &~$\textrm{Average~times~of~borrowing}$~                 &~$\textrm{Popularity}$~         \\ \hline

 ~7\#~    &~ $\textrm{Total~number~of~borrowers}$ ~        &~$\textrm{Average~amount~borrowed}$~        &~$\textrm{Total~turnover}$~         \\ \hline

 ~8\#~    &~ $\textrm{Loan~balance}$  ~                   &~$\textrm{Investment~dispersion}$~   &~$\textrm{Weighted~turnover}$~       \\ \hline

 ~9\#~    &~ $\textrm{Transaction~volume}$~            &~$\textrm{Loan~dispersion}$~       &~$\textrm{Transaction~volume}$~         \\ \hline

 ~10\#~   &~ $\textrm{Weighted~turnover}$ ~                 &~$\textrm{Total~number~of~borrowers}$~                  &~$\textrm{Average~times~of~borrowing}$~      \\ \hline

\end{tabular}
}\vspace{-0pt}
\end{table*}

Table~\ref{T:GraphInformation22} presents a comparison of the results obtained using the competing methods. For each method, we display the top 10 features in terms of relevance to the average annualized interest rate. It is worth noting that all three methods can identify some similar influential factors but differ from each other in the remaining factors. For instance, both InLasso and InElasticNet rank \textbf{the average amount invested}, \textbf{average times of investment}, and \textbf{online time} as the top three most influential factors. This is reasonable because a longer online time indicates that the P2P platform is in operation for a relatively longer period of time, and is less risky. Moreover, a larger average amount invested and a higher level of the average times of investment indicate a higher preference of the investors for the P2P lending platform due to a higher degree of security. In addition, both methods consider \textbf{investment dispersion} as a relevant feature but with different rankings, i.e., the 4th for InElasticNet and the 8th for InLasso. This is reasonable because investment dispersion is highly correlated with the average amount invested and the average times of investment. Therefore, when InElasticNet ranks these factors high, it also tends to rank investment dispersion high. This implies that our proposed method can encourage a grouping effect for highly correlated features. This is further demonstrated by the fact that \textbf{the rankings of popularity (6th)}, \textbf{total turnover(7th)}, and \textbf{weighted turnover (8th)} are close to each other. Moreover, these three factors are also correlated to each other. Whereas for InLasso, when groups of correlated features exist, it can only select one feature from the group. Therefore, InLasso may fail to recognize some highly relevant features.

When compared to ElasticNet, it is worth noting that although both Elastic Net and the proposed method (InElasticNet) can identify \textbf{online time}, \textbf{popularity}, \textbf{weighted turnover}, \textbf{transaction volume}, \textbf{average times of borrowing}, and \textbf{investment dispersion} as influential factors, their rankings are quite different. This meets our expectations because both ElasticNet and InElasticNet can promote a grouping effect. However, as InElasticNet utilizes the structural information between pairwise feature samples, the results obtained are more encouraging. For instance, ElasticNet ranks \textbf{loan dispersion (1st)} and \textbf{investment dispersion (2nd)} as the most influential factors whereas InElasticNet ranks \textbf{the average amount invested} as the highest and \textbf{the average times of investment} as the second highest. Unfortunately, a higher level of loan dispersion and investment dispersion does not necessarily correspond to a safer P2P platform with a lower annualized interest rate. By contrast, a larger average amount invested and a higher level of the average times of investment often indicate a higher preference of the investors for the P2P lending platform due to a higher degree of security and a lower level of annualized interest rate. These results demonstrate the effectiveness of the proposed method for identifying the most influential factors for credit risk of P2P lending platforms.

The healthcare dataset is collected from a well-known medical platform in China\footnote{See the website http://www.haodf.com/ for more details}, which presents evaluation of doctors. The dataset consists of 2363 doctors (i.e., 2363 samples). For each doctor, we collect 13 features including 1) patients, i.e., total number of patients treated, 2) title, i.e., title of the doctor, 3) grade, i.e.,recommended grade of the doctor provided by the website, 4) notes, i.e., total number of notes of thanks posted by the patients, 5) gifts, i.e., total number of gifts received, 6) outpatients, i.e., total number of outpatients of the doctor, 7) city, i.e., city of the doctor, 8)appointments, i.e., total number of appointments received from the patients, 9) visits, i.e., total number of visits of the doctor's personal website, 10) contribution value, i.e., contribution value of the doctor, 11) posts, total number of posts published by the patients about the doctor, 12) votes, i.e., total number of votes received from the patients for a doctor, and 13) registration fee for the doctor.

We use the proposed feature selection method to evaluate the doctors. The registration fee is treated as the continuous target and we aim to identify which features are the most informative ones with respect to the target. Like for the P2P lending analysis, we also compare the proposed feature selection method with two alternative methods including Elastic Net~\cite{ElasticNet} and Interacted Lasso~\cite{DBLP:journals/prl/ZhangTBXH17}.

\begin{table*}
\vspace{-0pt}
\centering {
\tiny
\caption{Comparison of three methods for the healthcare dataset}\label{T:GraphInformation2}
\vspace{0pt}

\begin{tabular}{|c||c||c||c|}

  \hline
 ~Ranking ~      ~ &~ElasticNet~        &~InLasso~                    & ~InElasticNet~  \\ \hline \hline
 ~1\#~    &~ $\textrm{City}$  ~         &~$\textrm{Title}$~           &~$\textrm{Title}$~      \\ \hline

 ~2\#~    &~  $\textrm{Title}$ ~        &~$\textrm{Grade}$~                &~$\textrm{City}$~     \\ \hline

 ~3\#~    &~ $\textrm{Grade}$ ~         &~$\textrm{City}$~                 &~$\textrm{Grade}$~     \\ \hline

 ~4\#~    &~ $\textrm{Notes}$  ~        &~$\textrm{Votes}$~                &~$\textrm{Votes}$~      \\ \hline

 ~5\#~    &~ $\textrm{Votes}$ ~         &~$\textrm{Contribution~value}$~   &~$\textrm{Outpatients}$~     \\ \hline

 ~6\#~    &~ $\textrm{Appointments}$~   &~$\textrm{Visits}$~          &~$\textrm{Contribution~value}$~         \\ \hline

\end{tabular}
}\vspace{-0pt}
\end{table*}

Table~\ref{T:GraphInformation2} presents a comparison of the results obtained using the three methods. For each method, we display the top 6 features in terms of relevance to the registration fee. It is worth noting that all three methods can identify \textbf{title of the doctor}, \textbf{city located} and \textbf{grade of the doctor} as the top three most influential factors. However, the rankings of these factors are different. For instance, both InLasso and InElasticNet rank \textbf{title of the doctor} the first, but ElasticNet ranks this factor as second. Compared with InLasso, our method considers \textbf{city} as the second most influential factor whereas InLasso ranks \textbf{Grade} as the second. We believe the results obtained via our method is more reasonable because a doctor with a higher title and in bigger cities are more expensive. Although grade is also relevant to the registration fee of the doctor, it is not an objective evaluation criteria. In addition, both InLasso and InElasticNet consider \textbf{the number of votes} as the fourth highest influential factor. This is also reasonable because a greater number of votes received from the patients indicates a higher reputation of the doctor. An interesting finding is that only our method can identify \textbf{outpatients} as the top ranking features whereas the two competing methods consider \textbf{appointments} and \textbf{visits} as the most influential features. We believe outpatients is a more relevant feature to the registration fee of the doctor because a higher number of outpatients often indicate that more patients are willing to pay more to be treated by the doctor. In addition, outpatients and votes are closely related to each other, and only the proposed method can select the highly correlated features.

\section{Conclusion}\label{s5}
The main goal of feature selection is to automatically identify a subset of the most informative features that is small in size but high in classification accuracy. To realize this goal, in this paper, we have developed a new structurally interacting elastic net feature selection method. The major contributions of this paper are threefold. First, the proposed method can encapsulate structural relationships between feature samples into the feature selection process by representing features as graphs and samples as graph vertices. Accordingly, the informativeness matrix obtained is used to construct an optimization model to identify the features with maximum relevancy and minimum redundancy to the target feature. Second, to remedy the information loss caused by using graph-based feature representations, we formulate the feature selection problem using an elastic net regression model and solve this model using ADMM. This allows us to a) incorporate information from the original feature space, b) reduce the number of features to a small size and c) promote grouping effects. The experimental results on real datasets show that our method outperforms several well-known feature selection methods.

We plan to extend our method in a number of ways. First, in this paper, the constructed feature graphs are complete weighted graphs. However, in real world applications, not all connections may be dominant and useful. In other words, the complete weighted graphs may contain some noise. Therefore, one may want to define a sparser graph. Second, in our previous work~\cite{DBLP:journals/pr/Bai0TH15}, we have developed a number of quantum Jensen-Shannon kernels using both the continuous-time and discrete-time quantum walks. It would be interesting to extend the proposed feature selection method using the classical Jensen Shannon divergence to that using its quantum counterpart. Finally, the proposed feature selection method only considers the relationships between pairwise features, i.e., it only evaluates the two-order relationships between features. Our future work will extend the proposed method into a high-order feature selection method by establishing higher order relationships between features.

%

\section*{Acknowledgments}

This work is supported by the National Natural Science Foundation of China (Grant noa. 61602535 and 61503422), the Open Project Program of the National Laboratory of Pattern Recognition (NLPR), and the program for innovation research in Central University of Finance and Economics.

\bibliography{cuibibfile2}

\end{document}